\documentclass[twoside,11pt]{article}
\usepackage{xcolor,graphicx,comment,hyperref,setspace,fullpage}
\usepackage{amsmath,amssymb,natbib,algorithm2e,subcaption,upgreek,authblk}

\newtheorem{proposition}{Proposition}

\newtheorem{theorem}{Theorem}
\newtheorem{assumption}{Assumption}
\newtheorem{remark}{Remark}

\hypersetup{
    colorlinks=true,
    linkcolor=blue,
    filecolor=magenta,      
    urlcolor=cyan,
    citecolor=blue,
}

\DeclareMathOperator*{\argmax}{argmax}
\DeclareMathOperator*{\argmin}{argmin}

\newcommand{\bs}{\boldsymbol}
\newcommand{\JG}[1]{{\color{green}JG: #1}}

\newcommand{\REV}[1]{{\color{black} #1}}

\allowdisplaybreaks

\title{Partially Monitored Online Binary Classification: Apple Tasting Revisited}

\begin{document}
\title{Apple Tasting Revisited: Bayesian Approaches to Partially Monitored Online Binary Classification}
\author[1]{James A. Grant\thanks{j.grant@lancaster.ac.uk; corresponding author}}
\author[1]{David S. Leslie\thanks{d.leslie@lancaster.ac.uk}}
\affil[1]{Department of Mathematics and Statistics, Lancaster University, UK}

\maketitle

\begin{abstract}%
We consider a variant of online binary classification where a learner sequentially assigns labels ($0$ or $1$) to items with unknown true class. If, but only if, the learner chooses label $1$ they immediately observe the true label of the item. The learner faces a trade-off between short-term classification accuracy and long-term information gain. This problem has previously been studied under the name of the `apple tasting' problem. We revisit this problem as a partial monitoring problem with side information, and focus on the case where item features are linked to true classes via a logistic regression model. Our principal contribution is a study of the performance of Thompson Sampling (TS) for this problem. Using recently developed information-theoretic tools, we show that TS achieves a Bayesian regret bound of an improved order to previous approaches. Further, we experimentally verify that efficient approximations to TS and Information Directed Sampling via P\'{o}lya-Gamma augmentation have superior empirical performance to existing methods. \end{abstract}

\textbf{Keywords:}
Partial Monitoring, Binary Classification, Thompson Sampling, Information-Directed Sampling, P\'{o}lya-Gamma Augmentation

\section{Introduction}

Binary classification is a fundamental problem in statistics, machine learning, and many applications. Its online version, wherein a learner iteratively guesses the classes of items and has their true classes revealed has also been well-studied  \citep[e.g.][]{angluin1988queries,littlestone1990mistake,cesa1996worst,ying2006online}. In this paper, we study a variant of the stochastic online binary classification where the true class is \emph{only} revealed for items guessed to belong to a particular fixed class, irrespective of whether that guess is correct. 

Such a problem has previously been studied under the name of the `Apple Tasting Problem', inspired by a toy problem of learning to visually identify bad or rotten apples in a packaging plant \citep{helmbold1992apple,HelmboldEtAl2000}. In this example, a learner considers apples one by one and makes a choice whether or not to taste the apple based on its appearance - apples are either good or bad and their quality can, to some extent, be determined by their appearance. The learner wishes to minimise the number of good apples tasted (as a tasted apple cannot be sold), and maximise the number of bad apples tasted (as this prevents bad produce going to market - the learner is willing to sacrifice their own taste buds for profit). The difficulty of the problem enters through the aspects that the learner does not know the function relating appearance to quality and only receives information on the quality of the apple if they taste it.

Apples aside, conceptually similar problems may arise in numerous settings. For instance, in quality control settings products arrive to a decision-maker sequentially. Products may be satisfactory or faulty, and the decision-maker can determine this by investigating the product, subject to a further cost. It may be prohibitively costly to investigate every product but it is also important to identify and remove faulty products, to avoid greater costs further down the line. To balance their costs the decision-maker must make sufficiently many investigations to learn the characteristics of faulty products.

As another example, companies monitoring credit card fraud sequentially observe transactions, a small proportion of which are fraud. There is effectively no cost to allowing a non-fraudulent transaction to proceed, but costs are associated with investigating a transaction (whether fraudulent or not) and with allowing a fraudulent transaction to proceed. The natural aim is to allow genuine transactions to continue unimpeded, and investigate the fraudulent transactions, but immediate information on the class of a transaction can only be gathered through investigation.

More broadly, problems of this flavour can arise in a range of online monitoring and supervised learning settings: such as filtering spam emails \citep{sculley2007practical}, classifying traffic on networks, and label prediction in many contexts. An understanding of optimal algorithms for these sequential classification tasks is therefore valuable to decision-makers in a wide range of applications. The next section introduces a mathematical framework for the apple tasting problem.

\subsection{Logistic Contextual Apple Tasting}

We consider an online binary classification task taking place over $T$ rounds. In each round $t \in [T] \equiv \{1,\dots,T\}$, an item arrives with associated feature vector $x_t \in \mathcal{X} \subset \mathbb{R}^d$. This item has a latent class $C_t \in \{0,1\}$, which is modelled as having a stochastic dependence on the feature vector $x_t$, and an unknown parameter vector $\theta^* \in \Theta \subset \mathbb{R}^d$. Note that $\theta^*$ does not vary with $t$, and that $\Theta$ is known. Throughout this paper, the distribution of the class in round $t$ is given as \begin{align*}
    C_t ~|~ x_t \sim \REV{Bernoulli}(\sigma(x_t^\top \theta^*)),
\end{align*} where  $\sigma$ denotes the logistic function\footnote{The nature of our work is such that analogous theoretical results should be readily available for other suitably smooth transforms, e.g. the probit function.}, such that $\sigma(z)=(1+e^{-z})^{-1}$, $z \in \mathbb{R}$. 

Also in round $t$, an agent, who we call `the learner', guesses the class of the item with feature vector $x_t$. The guessed class is represented by $A_t\in\{0,1\}$. If $A_t=1$, \REV{the true class $C_t$ is revealed.} If $A_t=0$, no feedback is received. It is known that if the true class is $C_t=0$, choosing $A_t=0$ incurs zero loss, and $A_t=1$ incurs a loss of one. If $C_t=1$, choosing $A_t=1$ incurs loss $l_{11}\geq 0$, which represents the cost of intervention, and $A_t=0$, incurs cost $l_{01}\geq l_{11}$.

The expected loss of an action $a\in\{0,1\}$ in round $t$ with respect to parameter $\theta$ is then given by the function $\mu_t:\{0,1\}\times\Theta\rightarrow \mathbb{R}$ defined as, \begin{equation}
    \mu_t(a,\theta) := \begin{cases} &l_{01}\sigma\left(x_t^\top\theta\right),\quad \quad \quad \quad \quad \quad ~~  a=0, \\ 
    &1 + (l_{11}-1)\sigma\left(x_t^\top\theta\right), \quad \quad ~ a=1. \end{cases} \label{eq::lossfun}
\end{equation} We define the optimal action in round $t$ w.r.t. a parameter $\theta \in \Theta$ as $$\alpha_t(\theta) \in \argmin_{a \in \{0,1\}} \mu_t(a,\theta),$$ and let $A^*_t = \alpha_t(\theta^*)$, i.e. $A^*_t$ is the action optimal with respect to the true parameter $\theta^*$ in round $t$.

We are interested in the expected performance of the learner, given they use a particular rule (or \emph{policy}) $\varphi$ to \REV{take actions - i.e. sequentially predict classes. Successful policies necessarily use information gained from previous actions, captured in a history\footnote{We use $\varsigma$ here to denote the sigma-algebra to avoid overloading notation. Notice, in particular that $\mathcal{H}_{t-1}$ includes the round $t$ feature vector $x_t$.}, $$\mathcal{H}_{t-1}:=\varsigma(A_1,x_1,C_1A_1,\dots,A_{t-1},x_{t-1},C_{t-1}A_{t-1},x_t), ~~t \in \mathbb{N}.$$ To admit randomised policies, we introduce a sequence of independent random variables $(Z_t)_{t\in\mathbb{N}}$, and consider $\varphi$ to be given by a sequence of $\mathcal{H}_t$-adapted sampling distributions $(\varphi_t)_{t\in\mathbb{N}}$, with $A_t$ being drawn from $\varphi_t$ and being measurable with respect to the sigma-algebra generated by $(\mathcal{H}_{t-1},Z_t)$.} The learner's performance is captured by the \emph{Bayesian regret} of the policy $\varphi$ in $T$ rounds, \begin{equation*}
    BR(T,\varphi) = \mathbb{E}_0\left(\sum_{t=1}^T \mu_t\left(A_t,\theta^*\right) - \mu_t\left(A^*_t,\theta^*\right) \right),
\end{equation*} where this expectation is taken with respect to a prior, $\pi_0$, on $\theta^*$ supported on $\Theta$ \REV{as well as over randomness in the classes and policy.} 

The Bayesian regret measures the difference between the expected loss of an oracle decision maker who knows the status distribution of each item and acts optimally with respect to this information, and the expected loss of the learner making decisions according to the policy $\varphi$. We will be interested in the scaling of the Bayesian regret with respect to $T$ and to the dimension $d$ of the context vectors. 

We will principally be interested in the performance of the Thompson Sampling (TS) policy. In round $t$, TS plays the action $\alpha_t(\theta_t)$ where $\theta_t$ is drawn from the posterior distribution $\pi_{t-1}=\pi_0(\cdot ~|~ \mathcal{H}_{t-1})$. 

The idea behind TS - of choosing actions according to the current posterior belief - dates back as far as \cite{Thompson1933}. However it is in the last decade that it has become popular as a general approach for sequential decision-making problems, and has been shown to achieve optimal or near-optimal scaling of regret across many settings including multi-armed bandits \citep{AgrawalGoyal2013}, contextual bandits \citep{MayEtAl2012,agrawal2013thompson},  structured bandits \citep{RussoVanRoy2016,GrantLeslie2020}, and a linear variant of finite partial monitoring \citep{tsuchiya2020analysis}.

\subsection{Partial Monitoring} To fully characterise our logistic apple tasting problem and its best achievable regret, it is useful to cast the problem as part of the broader partial monitoring (PM) framework. The apple tasting problem is one of the simplest examples of a PM problem. Specifically, our proposed model is an example of PM with side information, as considered in \cite{BartokSzepesvari2012}. 

It can also be shown that subject to a rescaling of the loss matrix \citep{AntosEtAl2013,lienert2013exploiting}, our framework coincides with a logistic contextual bandit \citep{FilippiEtAl2010} with only two actions in each round (one always being a zero vector). The present problem however is a very specific instance of this much broader framework, and a bespoke treatment of apple tasting is therefore useful as a complement to existing theory for the much more general setting \citep{DongEtAl2019,FauryEtAl2020}.

A general PM problem with side information is formalised as a tuple $\mathbf{G}=(\mathbf{L},\bs\Phi,\mathcal{F})$, where $\mathbf{L}\in\mathbb{R}^{N\times M}$ is a loss matrix, $\bs\Phi \in \Sigma^{N \times M}$ is a feedback matrix\footnote{Here $\Sigma$ is some known, finite alphabet.}, and $\mathcal{F}$ is a set of possible mappings from contexts to outcome distributions. In our case,  \begin{equation*}
    \mathbf{L} =  \left( \begin{matrix} 0 & l_{01} \\ 1 & l_{11} \end{matrix}\right) \text{ and } \bs\Phi =  \left( \begin{matrix} 0 & 0 \\ 1 & \REV{0} \end{matrix}\right),
\end{equation*} and $\mathcal{F}$ is the set of logistic functions parameterised by $\theta\in\Theta$. 

In each of a series of rounds $t\in[T]$, a context $x_t\in\mathcal{X}$ is drawn (not necessarily at random). The learner observes the context and then chooses an action $n_t\in [N]$. An outcome $m_t\in[M]$ is drawn from the distribution $f(x_t)$ and the learner receives loss $\mathbf{L}_{n_t,m_t}$ and observes feedback $\bs\Phi_{n_t,m_t}$. In our problem, $n_t$ corresponds to the choice of label, $m_t$ is the true class, and the losses and feedbacks are the corresponding entries of $\mathbf{L}$ and $\bs\Phi$.

The best achievable regret in PM problems is well understood in a minimax sense. In particular, for both stochastic \citep{BartokEtAl2011,BartokEtAl2014} and adversarial \citep{BartokEtAl2010,AntosEtAl2013} finite PM (where $M$ and $N$ are finite), it has been shown that all games can be classified as either `trivial', `easy', `hard', or `hopeless' based on $(\mathbf{L},\bs\Phi)$, and have associated minimax regret of order $\Theta(1)$, $\Theta(\sqrt{T})$, $\Theta(T^{2/3})$, or $\Theta(T)$ respectively. \REV{This classification was initially developed in the non-contextual regime, but has since been shown to carry to linear contextual games with finite-actions \citep{KirschnerEtAl2020,kirschner2023linear}. The impact of context dimension and the number of actions on the best achievable regret is discussed further in the following sub-section.}

In all \REV{aforementioned settings (adversarial and stochastic, contextual and non-contexutal)}, the apple tasting problem is shown to belong to the class of `easy' problems, with $\Theta(\sqrt{T})$ minimax regret\footnote{Note that these bounds are in the frequentist setting, but such bounds automatically imply equivalent results in the Bayesian setting (the converse is not true). See e.g. Section 34.7 of \cite{lattimore2020bandit}}. It is worth noting that the term \emph{easy} refers to the learnability of the problem, but does not imply that the design of an optimal algorithm is trivial. 

For a general family of easy problems with side information (including apple tasting) \cite{BartokSzepesvari2012} prove that an upper-confidence-bound-based algorithm, CBP-SIDE, realises this $\Theta(\sqrt{T})$ regret. Furthermore, \cite{lienert2013exploiting} shows empirically that both CBP-SIDE and the LinUCB algorithm \citep{li2010contextual,chu2011contextual} for contextual bandits, are effective for a linear (as opposed to logistic) variant of our problem. 

However, in practice, upper-confidence-bound based approaches can be overly conservative, and only behave competitively when $T$ is very large. In this paper we show that an improved empirical performance can be achieved by randomised, Bayesian algorithms such as TS, without sacrificing theoretical guarantees.

\subsection{Related Literature}

As mentioned previously, our principal focus in this paper is the performance of TS - \cite{RussoEtAl2018} gives a detailed summary of developments around TS across many problems. Our analysis of the Bayesian regret is based on a recent line of information-theoretic analysis \citep{RussoVanRoy2016,DongVanRoy2018,DongEtAl2019,LattimoreSzepesvari2019} which has been shown to be useful for problems with large context or action sets. In utilising these ideas in the apple tasting setting, we extend a number of existing results for finite parameter sets to compact $\Theta \subset \mathbb{R}^d$. An alternative strategy for bounding the Bayesian regret, not considered here, has been to exploit frequentist confidence sets to construct high probability guarantees on which actions are selected \citep{russo2014learning,grant2019adaptive,GrantLeslie2020}.

A related algorithm, inspired by the link between information gain and the necessary exploration in sequential decision-making problems is Information-Directed Sampling (IDS), introduced in \cite{RussoVanRoy2014,RussoVanRoy2018}. Like TS, IDS also selects at random based on the posterior belief, but constructs the distribution from which this sample is drawn based on a trade-off of expected regret, and expected information gain given the feedback on the upcoming action. IDS (and frequentist approximations thereof) has been applied to bandit, partial monitoring, and reinforcement learning problems \citep{liu2017information,KirschnerKrause2018,KirschnerEtAl2020,kirschner2020asymptotically,arumugam2020randomized} and shown strong empirical and theoretical results, comparable to those for TS. We discuss the use of IDS for apple tasting in Section \ref{sec::altapp}, and evaluate it empirically alongside TS in Section \ref{sec::experiments}. 

\REV{Due to the previously mentioned connection between apple tasting and logistic bandits, state-of-the-art algorithms for those problems are also worth mentioning. Beyond TS approaches, much of the literature \citep{FauryEtAl2020,abeille2020instance,faury2022jointly,lee2023improved,zhang2024online} has focussed on deriving sharp concentration results and confidence sets for UCB-type policies, however these typically target settings with larger action sets and would not consider the particular information structure of apple tasting. \cite{jun2021improved} propose SupLogistic, which combines optimism and data bucketing, following the ideas of SupLinRel \citep{auer2002using}, and $O(\sqrt{d})$ dependence on $d$ for problems with finitely many arms per round, but we see in Section \ref{sec::experiments} that it has an inferior empirical performance to TS and IDS approaches. \cite{mason2022experimental} propose an algorithm with similarly sharp dependence on $d$. However it assumes the set of arms is fixed across all rounds.}

The present paper is the first work we aware of that specifically applies TS to apple tasting, but previous work has considered its use for logistic bandits. For logistic contextual bandits, the implementation of exact TS (i.e. the policy that draws its sample from the exact posterior) is infeasible due to the intractability of the posterior distribution. It is therefore necessary to sample from an approximation of the posterior to implement a TS-like approach. \cite{DumitrascuEtAl2018} recently proposed an approximation based on Polya-Gamma augmentation \citep{PolsonEtAl2013,WindleEtAl2014} which has improved convergence properties over Laplace approximation originally used by \cite{chapelle2011empirical}. We investigate such a Polya-Gamma augmentation-based approximation in the context of apple tasting in Section \ref{sec::PGTS}. The effect of approximation of the posterior on the performance of TS is an area of increasing interest, as \cite{PhanEtAl2019} have recently proved that a small constant approximation error can induce linear regret in the application of TS to certain simple multi-armed bandit problems. Appropriately designed approximate algorithms can be successful however, as shown theoretically \citep{mazumdar2020thompson} for particular Langevin approximation algorithms, and empirically in a range of settings \citep[e.g.][]{urteaga2018variational}.


The apple tasting problem is not the only variant of online classification where labels are not revealed in every round. In selective classification (or classification with a reject (or abstention) option) \citep[e.g.][]{chow1957optimum,sayedi2010trading,wiener2011agnostic}  the learner may decline to label items, thus mitigating the risk of labelling when they have high uncertainty.  Conversely, in classification with selective sampling \citep{cesa2009robust,orabona2011better,cavallanti2011learning,dekel2012selective,agarwal2013selective}, the learner must label all items, but observing labels is costly, and the learner has the option to decline to observe the label if it is deemed to have insufficient informational value. The same problem has also been studied under the name `online active binary classification' \citep{monteleoni2007practical,liu2015efficient}. Both of these variants differ from apple tasting in that they have a more complex action set.

\cite{gentile2014multilabel} consider a multi-class label prediction problem where the learner chooses a subset of possible labels, in each round, and only observes true labels if they are part of their subset. While this also lies at the intersection of classification and partial monitoring, when the number of classes is reduced to two, i.e. in the binary classification setting, this problem reduces to the usual full-feedback online problem. Apple tasting is therefore more challenging in the 2-class setting due to the information imbalance between the actions.

\subsection{Motivations and Contributions}

Our motivations for a renewed treatment of apple tasting are threefold. Firstly, despite the existence of alternative theoretically justified approaches, new developments in the theoretical understanding of TS allow us to derive guarantees for the empirically superior TS policy. Second, the apple tasting setting strikes a sufficient balance between simplicity and complexity to allow an uncluttered study of the effect of zero-information actions in PM, and of posterior approximation to the performance of TS. Finally, as outlined in the first section, the range of applications of the apple tasting problem are broad, and our empirical investigation of methods such as TS that have been popularised with the last decade is therefore useful and pertinent.

The principal contributions of our work are the following: \begin{itemize}
    \item We provide an information-theoretic analysis of the Bayesian regret of TS for logistic contextual apple tasting (LCAT). This gives rise to an $\tilde{O}(\sqrt{dT})$ bound on regret which is optimal with respect to horizon $T$ (up to logarithmic factors), and sharper with respect to the feature dimension $d$ than  frequentist bounds for \REV{state-of-the-art} UCB-type algorithms. Notably, the bound is also of an improved order with respect to $d$ than the $\tilde{O}(d\sqrt{T})$ bounds achievable \REV{for most algorithms, including TS,} in the more general contextual bandit setting, \REV{and the empirical performance is greatly superior to the one algorithm (SupLogistic) with a matching regret bound}. Our analysis extends theoretical techniques previously only used for finite parameter spaces to the more readily modelled setting of compact but infinite parameter spaces.
    \item We adapt the P\'{o}lya-Gamma TS scheme of \cite{DumitrascuEtAl2018} to give approximate TS and information directed sampling (IDS) schemes appropriate to LCAT. In the TS setting, we show that this scheme is an asymptotically consistent approximation to exact TS.
    \item We identify a potential issue in the application of IDS to contextual problems, or those with non-stationary expected information gains. Namely, that it may fail to take account of the magnitude of the information gain and make counter-intuitive selections as a result. We explain how a tunable variant avoids this issue.
    \item We validate the efficacy of the P\'{o}lya-Gamma-based TS and tunable-IDS by showing their superior performance to competitor approaches on simulated data.
\end{itemize}

Our theoretical results on TS are given in Section \ref{sec::theory}. In Section \ref{sec::PGTS} we discuss the Polya-Gamma augmentation scheme necessary for a practical implementation of TS, and in Section \ref{sec::altapp} we discuss the limitations of a similar implementation of IDS. Finally, we demonstrate the efficacy of TS numerically in Section \ref{sec::experiments}.

\section{Bayesian Regret of Thompson Sampling} \label{sec::theory}

Our first theoretical result is given in this section. It bounds the Bayesian regret of TS under any sequence of bounded feature vectors. To complete our formalisation of the setting in which theoretical guarantees can be established, we suppose w.l.o.g. that the parameter set $\Theta$ lies in the \REV{Euclidean} unit ball in $\mathbb{R}^d$, i.e. $\Theta \subset B^d_1$, and that there exists a bound $x_{\max} < \infty$ on the dimensions of feature vectors, i.e. $||x||_\infty\leq x_{\max}$, for all $x\in \mathcal{X}$. \REV{Finally, we make the following assumption on the combination of the prior and sequence of contexts.} \begin{assumption} \label{assumpt:1}\textbf{- \emph{Frequent Informative Actions}}
\REV{The sequence of contexts $\{x_t\}_{t=1}^T$ and prior distribution $\pi_0$ are such that the expected number of rounds between selections of the informative action under TS is bounded above by a constant, i.e. if the round in which the informative action is used for the $u^{th}$ time is defined as  $$N(u):=\min_{t\in\mathbb{N}}\left\{t: \sum_{s=1}^t \mathbb{I}\{A_s=1\}=u\right\}$$ with $N(0):=0$, there exists $C>0$ such that for all $u>1$,} \begin{equation}
\REV{\mathbb{E}_0\left(N(u)-N(u-1)\right) < C.}
\end{equation}
\end{assumption} \REV{If the probability of choosing the informative action was bounded below by $1/C$ in every round, Assumption 1 would follow immediately. In practice, Assumption 1 may be satisfied when the prior has sufficiently heavy tails or the contexts are sufficiently variable and uncorrelated. The latter may be interpreted as a reasonable condition to impose on the problem to suppose learning is achievable. We may now state our Bayesian regret bound.}

\begin{theorem} \label{thm::TSBR}
For the contextual logistic apple tasting problem instantiated by $\theta^*\sim\pi_0$, the Bayesian regret of the Thompson Sampling policy, $\varphi^{TS}$, in $T$ rounds satisfies, \begin{equation}
    BR\left(T,\varphi^{TS}\right)=O\left(\sqrt{dT\log(T)}\right). \label{eq::mainBR}
\end{equation}
\end{theorem} In the following section, \ref{sec::preliminaries}, we introduce some information theoretic concepts which are required in the proof of the bound, which is given in Section \ref{sec::mainproof}.

The result in Theorem 1 may imply the stronger performance of TS than alternative approaches. The {CBP-SIDE} algorithm of \cite{BartokSzepesvari2012} achieves  a near-optimal frequentist regret, but still with an order greater than our Bayesian regret bound: \begin{equation*}
    Reg\left(T,\varphi^{CBP-SIDE}\right)=O\left(d^2\log(T)\sqrt{T}\right).
\end{equation*} In particular, its dependence on the dimension of the parameter vector is worse by a factor of $d^{3/2}$. Casting the problem as a contextual logistic bandit, as outlined in \cite{AntosEtAl2013,lienert2013exploiting} and directly utilising the existing results of \cite{DongEtAl2019} for said more general setting gives an $\tilde{O}(d\sqrt{T})$ bound on Bayesian regret. Our difference of a factor of $\sqrt{d}$ in \eqref{eq::mainBR} is a result of our bespoke analysis for the 2-action apple tasting setting. \REV{\cite{jun2021improved} show an $\tilde{O}(\sqrt{dT})$ for their SupLogistic algorithm, in a logistic contextual bandit with finitely many arms, and this is the closest result to ours, however we will show in Section \ref{sec::experiments} that its empirical performance is considerably worse than implementations of TS and IDS, on LCAT problems.}

The other previously existing approaches for contextual apple tasting of \cite{HelmboldEtAl2000}, which transform binary classification algorithms to apple tasting algorithms, guarantee a regret that is sublinear in $T$ but that is linear in the size of the context set $|\mathcal{X}|$. Such bounds are therefore not useful in the present setting with infinitely many possible contexts. 

\REV{A bespoke lower bound on the regret for LCAT (incorporating feature dimension) has not yet been derived, although we conjecture that Theorem \ref{thm::TSBR} matches the best performance up to logarithmic terms. It may appear that our result contradicts any extension of the well known $\Omega(d\sqrt{T})$ bound for linear contextual bandits in \cite{dani2008stochastic}, however their bound applies to a problem with continuous action sets\footnote{\REV{Specifically, their $d$ is constrained to be even, and the action set is comprised of a Cartesian product of $d/2$ circles}}. The $\Omega(\sqrt{dT})$ bound in \cite{chu2011contextual} applies to stochastic linear bandits with finitely many arms (per round) and it is an extension of this result that is likely to apply to LCAT, where there are two actions per round.}

\subsection{Preliminaries} \label{sec::preliminaries}

Before giving the proof of Theorem \ref{thm::TSBR}, we require some additional notation and concepts. Firstly, in-keeping with the notation for more general PM problems, we define the incurred loss and observed signal as $L_t=L(C_t,A_t)$ and $\Phi(A_t)=\REV{C_t A_t}$.

Our bound will rely on information-theoretic techniques, and as such a definition of the mutual information between probability distributions is necessary. For random variables $X$, and $Y$, following distributions $P_X$, and $P_Y$ respectively, \REV{with joint distribution $P_{X,Y}$} define the mutual information between $X$ and $Y$ as \begin{equation*}
    I(X;Y) := D_{KL}(P_{X,Y}||P_X \otimes P_Y),
\end{equation*} where $D_{KL}$ denotes the Kullback-Leibler divergence\footnote{\REV{Defined, for completeness, between two probability measures $P$ and $P'$ on a common measurable space with $P$ absolutely continuous with respect to $P'$ as $D_{KL}(P||P')=\int\log\left(\mathrm{d}P/\mathrm{d}Q\right)\mathrm{d}P$.}}. \REV{For a triple of variables, where $Z$ is introduced and follows $P_Z$, we may write \begin{equation*}
    I(X;(Y,Z)) := D_{KL}(P_{X,Y,Z} || P_X \otimes P_{Y,Z}).
\end{equation*}}

Related to this, a key quantity will be the expected information gained by the learner about the parameter $\theta^*$ in a single round. To define this, let $\pi_{t}$ be the posterior \REV{probability distribution on $\theta^*$ induced by the combination of prior $\pi_0$ and the likelihood of history} $\mathcal{H}_t$, and introduce $I_{t+1}$ as a function giving the mutual information between random variables under this posterior.
%
%
For random variables $X, Y$ and \REV{$Z$ and posterior $\pi_{t-1}$} define \begin{align*}
    I_t(X;Y)&= \REV{D_{KL}(\pi_{t-1}(X,Y)~||~\pi_{t-1}(X) \otimes \pi_{t-1}(Y))}, \\
    \REV{I_t(X; (Y,Z))}&=\REV{D_{KL}(\pi_{t-1}(X,Y,Z)~||~\pi_{t-1}(X) \otimes \pi_{t-1}(Y,Z)).}
\end{align*} 
The expected information gained about $\theta^*$ in a single round $t$ by the learner using TS is then represented by  
$I_t(\theta^*; (\theta_t,\Phi_t(\alpha_t(\theta_t))))$. 



This expected information gain plays a key role in the following quantity called the \emph{information ratio}. \REV{Let $\mathbb{E}_{t}$ denote expectation with respect to $\pi_{t}$.} For any $\Theta$-valued random variables $\theta, \theta'$ and $t \in [T]$ we define the information ratio as \begin{equation}
    \Gamma_t(\theta,\theta') = \frac{\left[\mathbb{E}_{t-1}\left( \mu_t\left(\alpha_t(\theta'),\theta^*\right)-\mu_t\left(\alpha_t(\theta),\theta^*\right) \right)\right]^2}{I_t\left(\theta; (\theta',\Phi_t(\alpha_t(\theta'))) \right)}. \label{eq::infratio}
\end{equation}  When $\theta=\theta^*$ this is the ratio of the square of the expected regret incurred in round $t$ by a decision-maker acting as though $\theta'$ is the true parameter and the expected information gained about $\theta^*$ as a result of their decision. Thus a large information ratio corresponds to high regret and low expected information, whereas a small information ratio corresponds to low regret and high expected information.

The information ratio is a quantity, introduced in a more general setting by \cite{RussoVanRoy2016}, which allows for a useful decomposition of the Bayesian regret of Thompson Sampling and related randomised approaches. When the information ratio can be uniformly bounded for all $t \in [T]$ several studies have successfully derived order-optimal bounds on the Bayesian regret in various settings \citep{RussoVanRoy2016,RussoVanRoy2018,DongEtAl2019,LattimoreSzepesvari2019}. In such settings, the bound is expressed terms of such a uniform bound on $\Gamma_t(\cdot,\cdot)$, and \REV{$H(\theta^*):=\mathbb{E}_0(-\log(\theta^*))$}, the \REV{Differential} entropy of the distribution on $\theta^*$, \REV{the latter of which, informally, upper bounds the information to be learned about $\theta^*$.}

Here however, the realisation of such a bound would not be finite as the parameter space $\Theta$ is not discrete, and thus $H(\theta^*)$ lacks a finite bound. Fortunately, this does not mean that the problem of bounding the regret in our setting is hopeless. To achieve sublinear regret it is not necessary to learn the distribution over all of $\Theta$. Following \cite{DongVanRoy2018} we introduce a \emph{rate distortion} of the parameter $\theta^*$ to learn a simpler $\epsilon$-optimal action function rather than the exact optimal action function.

For $\theta,\theta' \in \Theta$, and a round $t \in [T]$ we define the \emph{distortion rate} as \begin{equation*}
    d_t(\theta,\theta') = \mu_t(\alpha_t(\theta),\theta') - \mu_t(\alpha_t(\theta'),\theta').
\end{equation*} This measures the difference in the expected loss computed with respect to $\theta'$ of the optimal action given $\theta$ and $\theta'$. It happens to coincide with the regret of TS when $\theta'=\theta^*$, the true parameter, and $\theta=\theta_t$, the sample drawn by TS in round $t$. Further, define $\{\Theta_{k} \}_{k=1}^K$ to be a partition of $\Theta$ in $K$ parts such that, for a fixed and arbitrary $\epsilon>0$ \begin{equation*}
    d_t(\theta,\theta') \leq \epsilon, \quad \forall \theta, \theta' \in \Theta_{k}, ~\forall t\in[T], \enspace \text{ for any } k \in [K].
\end{equation*} \REV{Note that the existence of such partitions (which are used only as tools in the theoretical analysis) is verified later as a part of Proposition \ref{prop::partitionsize}. The analysis itself is agnostic to the \emph{particular} choice of partition, requiring only that it is chosen to satisfy the above condition - which could be done with knowledge of $\mathcal{X}$ only, rather than $\{x_t\}_{t\in[T]}$.} We then define an associated indexing random variable $\phi_\epsilon$ on $[K]$ such that \begin{equation}
    \phi_\epsilon=k \Leftrightarrow \theta^* \in \Theta_k. \label{eq::phiepsdef}
\end{equation}
 This variable indicates which cell of the partition the true parameter $\theta^*$ lies in. The entropy of $\phi_\epsilon$, $H(\phi_\epsilon)$ is \REV{upper} bounded by $\log(K)$.  Thus if the structure of $\Theta$ permits a small $K$, $H(\phi_\epsilon)$ can be much smaller than $H(\theta^*)$. In what follows, we will derive a bound on the Bayesian regret of TS that depends on $H(\phi_\epsilon)$.

\subsection{Proof of Theorem \ref{thm::TSBR}} \label{sec::mainproof}

The regret upper bound arises as a result of approximating the regret of TS using the posterior on $\theta^*$ with the regret of TS using a discrete distribution. The basic premise is to relate the regret to that of an algorithm that is expected to incur at least $o(\epsilon)$ regret in each round due to discretisation, and then study the additional regret incurred in the simpler discrete setting. The bound of the desired order is then obtained by specifying $\epsilon$ as a decreasing function of $T$. We emphasise that $\epsilon$ and the discretised variant of TS are purely hypothetical constructs for the proof, and are not required for the actual implementation of TS - thus the decision-maker does not have to worry about identifying and tuning them.

The first step in our proof constructs a series of discrete random variables $\tilde\theta^*_t$, $t\in[T]$ which approximate $\theta^*$. These random variables are functions of $\theta^*$ whose realisations lie in the same cell of the partition as $\theta^*$. However, their dependence on $\theta^*$ can be expressed entirely via the variable $\phi_\epsilon$, such that $\tilde\theta^*_t$ is independent of $\theta^*$ conditioned on $\phi_\epsilon$. The following proposition asserts the existence of these random variables, and verifies their certain key properties. This result \REV{adapts} a similar result (Proposition 2 of \cite{DongVanRoy2018}) to \REV{the present setting, extending a) to handle the fact that $\Theta$ is not a discrete set, and b) to consider the posterior mass placed on parameters corresponding to optimality of the informative action, a key aspect of the apple tasting analysis.} Below, the random variable $\tilde\theta_t$, which has the same marginal distribution as $\tilde\theta^*_t$, can be thought of as a discretised analog of the Thompson sample $\theta_t$.

\begin{proposition} \label{prop::thetatilde}
For $\phi_\epsilon$ as defined in \eqref{eq::phiepsdef}, there exists a \REV{$\tilde\pi_t$-distributed} random variable $\tilde\theta^*_t$, supported on up to $2K$ points in $\Theta$, in each round $t$ satisfying the following properties: \begin{enumerate}
    \item[(i)] Conditioned on $\phi_\epsilon$, $\tilde\theta^*_t$ is independent of $\theta^*$,
    \item[(ii)] $\mathbb{E}_{t-1}\left(\mu_t\left(\alpha_t(\theta_t),\theta^*\right)-\mu_t\left(\alpha_t(\theta^*),\theta^*\right) \right) \leq \epsilon + \mathbb{E}_{t-1}\left(\mu_t\left(\alpha_t(\tilde\theta_t),\theta^*\right)-\mu_t\left(\alpha_t(\tilde\theta^*_t),\theta^*\right)\right)$, a.s.,
    \item[(iii)] $\REV{I_{t}}\left(\phi_\epsilon; (\tilde\theta_t, \Phi(\alpha_t(\tilde\theta_t))) \right) \leq \REV{I_{t}}\left(\phi_\epsilon; (\theta_t, \Phi(\alpha_t(\theta_t))) \right)$, a.s.,
    \REV{\item[(iv)] $\tilde\pi_{t-1}(\tilde\Theta_t) = \pi_{t-1}(\tilde\Theta_t)$.}
\end{enumerate} where  $\tilde\theta_t$ is independent from and identically distributed to $\tilde\theta^*_t$, \REV{and $\tilde\Theta_t:=\{\theta\in \Theta: \alpha_t(\theta)=1\}$.}
\end{proposition}   Properties (ii) and (iii) say that the distribution of $\tilde\theta_t$ is such that the extra regret incurred by following TS using $\tilde\theta_t$ is no more than $\epsilon$, and that the information gain about the compressed random variable $\phi_\epsilon$ is not more than that gained using TS. The proof of Proposition \ref{prop::thetatilde} is given in Appendix \ref{app::thetatilde}.

We use the properties of $\tilde\theta_t^*$, and $\Gamma_t$ to decompose the expected regret in a single round. Essentially, the steps below bound per-round-regret by accepting a constant regret of $\epsilon$, to move from considering regret of TS to the regret of the discretised TS. For the regret in round $t$, $\Delta_t=\mu\left(\alpha_t(\theta_t),\theta^*\right)-\mu\left(\alpha_t(\theta^*),\theta^*\right)$, we have, \begin{alignat}{2}
    \mathbb{E}_{t-1}(\Delta_t) &= \mathbb{E}_{t-1}\left(\mu\left(\alpha_t(\theta_t),\theta^*\right)-\mu\left(\alpha_t(\theta^*),\theta^*\right)  \right) && \nonumber \\
    &\leq \epsilon + \mathbb{E}_{t-1}\left(\mu_t\left(\alpha_t(\tilde\theta_t),\theta^*\right)-\mu_t\left(\alpha_t(\tilde\theta^*_t),\theta^*\right)  \right) \quad    &&\text{by (ii)} \nonumber \\
    &= \epsilon + \sqrt{\Gamma_t(\tilde\theta^*_t,\tilde\theta_t)I_t\left(\tilde\theta^*_t;\left(\tilde\theta_t,\Phi(\alpha_t(\tilde\theta_t))\right)\right)} &&\text{by definition of } \Gamma_t \nonumber \\
    &\leq \epsilon + \sqrt{\Gamma_t(\tilde\theta^*_t,\tilde\theta_t)I_t\left(\phi_\epsilon;\left(\tilde\theta_t,\Phi(\alpha_t(\tilde\theta_t))\right)\right)} && \label{eq::DPI} \\
    &\leq \epsilon + \sqrt{\Gamma_t(\tilde\theta^*_t, \tilde\theta_t)I_t\left(\phi_\epsilon;\left(\theta_t,\Phi(\alpha_t(\theta_t))\right)\right)} &&\text{by (iii).} \label{eq::endofinit}
\end{alignat} \REV{The inequality \eqref{eq::DPI} uses the data processing inequality, valid since $\tilde\theta_t^*~|~x_t$ is (by the construction in the proof of Proposition 2) a deterministic function of $\phi_\epsilon~|~x_t$.} The consequence of \eqref{eq::endofinit} is that the regret in a single round has been decomposed in terms of the information ratio, and expected information gain. We proceed to bound this further, via a uniform bound on the information ratio, as given by the following proposition.

\begin{proposition} \label{prop::infratiobound}
There exists a constant \REV{$G>0$} such that for every round $t\in\mathbb{N}$  \begin{equation}
    \Gamma_t(\tilde\theta^*_t, \tilde\theta_t) \leq \frac{\REV{G}}{\REV{\tilde\pi_{t-1}(\tilde\Theta_t)}}, \label{eq::infratiobound}
\end{equation} where $\REV{\tilde\Theta_t:=\{\tilde\theta\in\tilde\Theta:\alpha_t(\tilde\theta)=1\}}$ 
is the set of parameters \REV{in the support of $\tilde\theta_t$} such that the revealing label is chosen, and  $\REV{\tilde\pi_{t-1}(\tilde\Theta_t)=\sum_{\tilde\theta\in\tilde\Theta_t}\tilde\pi_{t-1}(\tilde\theta)}$ is the posterior mass placed on this set. 
\end{proposition} Similar to Proposition \ref{prop::thetatilde}, the proof of this Proposition extends ideas from \cite{DongVanRoy2018} to non-finite $\Theta$, and the partial monitoring setting. \REV{Here, our finite action set (per round) allows a bound without dependence on $d$, in contrast to \cite{DongVanRoy2018}, where the larger action set leads to an $O(d)$ bound.} The techniques used in this extension could also be used to extend to non-finite parameter spaces in other problems. Principally, the proof consists of lower bounding the information gain via Pinsker's inequality, and relating the expected loss function to the sigmoid function to bound the information ratio. The full proof is provided in Appendix \ref{app::infratiobound}.

It follows from Proposition \ref{prop::infratiobound} that the per-round regret can be bounded as follows, \begin{alignat}{2}
    \mathbb{E}_{t-1}(\Delta_t) &\leq \epsilon + \sqrt{\frac{\REV{G}}{\REV{\tilde\pi_{t-1}(\tilde\Theta_t)}} I_t\left(\phi_\epsilon;  (\theta_t,\Phi_t(\alpha(\theta_t)))\right)} &&\text{by \eqref{eq::endofinit} and \eqref{eq::infratiobound}} \nonumber \\
    &= \epsilon + \sqrt{\frac{\REV{G}}{\REV{\tilde\pi_{t-1}(\tilde\Theta_t)}} \pi_{t-1}(\Theta_t)I_t\left(\phi_\epsilon; (\theta_t,\Phi_t(1))\right)} \quad \quad  && \nonumber \\
    &= \epsilon + \sqrt{\REV{G} I_t\left(\phi_\epsilon; (\theta_t,\Phi_t(1))\right)}, &&\REV{\text{Prop \ref{prop::thetatilde}, Part (iv)} }\label{eq::simplerregret}
\end{alignat} where the first equality is true because if $A_t=0$, there is no information gain. Now, aggregating the regret over $T$ rounds, we have \begin{alignat}{2}
    BR(T,\REV{\varphi^{TS}}) &= \sum_{t=1}^T\REV{\mathbb{E}_0}\left(\mathbb{E}_{t-1}(\Delta_t)\right) &&\text{Tower Rule} \nonumber \\
    &\leq T\epsilon + \sum_{t=1}^T \REV{\mathbb{E}_0}\left(\sqrt{\REV{G} I_t\left(\phi_\epsilon; (\theta_t,\Phi_t(1))\right)}\right) &&\text{by \eqref{eq::simplerregret}}\nonumber \\
          &\leq T\epsilon + \sqrt{\REV{G}T \sum_{t=1}^T \REV{\mathbb{E}_0}\left(I_t\left(\phi_\epsilon; (\theta_t,\Phi_t(1))\right)\right)} \quad \quad && \label{eq::sumentropy}
\end{alignat} \REV{with the final step due to the Cauchy-Schwarz inequality.} 

\REV{Handling the summation in \eqref{eq::sumentropy} is more complex than in settings where all actions provide an information gain. In such settings, the corresponding sum is readily upper bounded by the entropy of $\phi_\epsilon$ \citep{DongVanRoy2018}. Here, we still obtain a constant expression, but it is larger, and we must invoke Assumption 1 to achieve this, since we require some guarantee that the informative action is played sufficiently often. 

To proceed we introduce notation $S(t)=\sum_{v=1}^t \mathbb{I}\{A_v=1\}$ to count the number of rounds up to and including round $t$ in which class 1 is selected. The summation in \eqref{eq::sumentropy} may then be decomposed as follows,} \begin{align}
   \REV{\sum_{t=1}^T\mathbb{E}_0\left(I_t\left(\phi_\epsilon; (\theta_t,\Phi_t(1))\right)\right)} &= \REV{\sum_{t=1}^T \sum_{u=0}^{t-1} \mathbb{E}_0\left(I_t\left(\phi_\epsilon; (\theta_t,\Phi_t(1))\right)|S(t-1)=u\right)\mathbb{P}_0\left(S(t-1)=u\right)} \nonumber \\
   &\leq \REV{C\sum_{t=1}^T \mathbb{E}_0\left(I_t\left(\phi_\epsilon; (\theta_t,\Phi_t(1))\right)|S(t-1)=t-1\right)} \nonumber \\
   &\leq \REV{C\cdot I(\phi_\epsilon,\theta^*)} \nonumber \\
   &\leq \REV{C \cdot H(\phi_\epsilon)}. \label{eq::informationbound}
\end{align}  \REV{with the first inequality using Assumption \ref{assumpt:1}} and the final inequality holding since the expected information gained about $\phi_\epsilon$ is ultimately bounded by its entropy, i.e. \begin{equation*}
    \sum_{t=1}^T I_t\left(\phi_\epsilon; (\theta_t,\Phi_t(1))\right) \leq I(\phi_\epsilon;\theta^*)\leq H(\phi_\epsilon).
\end{equation*}

\REV{Combining \eqref{eq::sumentropy} and \eqref{eq::informationbound}, we bound the Bayesian regret as,}
\begin{align}
    \REV{BR(T,\varphi^{TS})}&\REV{\leq T\epsilon+ \sqrt{TCG H(\phi_\epsilon)}.} \label{eq::decomposed}
\end{align} The final step in the proof is to bound the $H(\phi_\epsilon)$ term in (\ref{eq::decomposed}). We recall that $H(\phi_\epsilon) \leq \log(K)$, where $K$ is the size of the partition of $\Theta$. The following proposition bounds $K$, and has its proof in Appendix \ref{app::partitionsize}, \REV{for convenience in its presentation we define $l_{max}:=\max(l_{01},1-l_{11})$}.

\begin{proposition} \label{prop::partitionsize}
There exists a partition $\{\Theta_k\}_{k=1}^K$ of $\Theta$, satisfying \begin{equation}
    \label{eq::partcond}
    d_t(\theta,\theta')\leq \epsilon, ~ \theta,\theta' \in \Theta_k,~ \forall k \in [K], ~\forall t \in [T],
\end{equation} for any \REV{$\epsilon \in (0,l_{max})$}, such that \begin{equation}
    \label{eq::Ksizebound}
    K \leq  \left(\frac{3l_{max}x_{max}}{\epsilon}  \right)^d.
\end{equation}
\end{proposition} 
The proof is completed by combining \eqref{eq::decomposed} and \eqref{eq::Ksizebound} and choosing $\epsilon=O\left(\sqrt{CGd/T}\right)$, \REV{to realise the bound} \begin{align*}
    \REV{BR(T,\phi^{TS}) = O\left(\sqrt{dCGT\log\left(l_{max}x_{max}T\right)}\right).} \quad \square
\end{align*}

\section{P\'{o}lya-Gamma Thompson Sampling} \label{sec::PGTS}

The logistic classification model is such that the posterior on $\theta^*$, $\pi_t$, for any $t\geq 1$ is intractable. This renders the implementation of TS via samples from the exact posterior in each round infeasible, and necessitates the use of samples from an approximate posterior. 

In the related setting of the logistic contextual bandit, \cite{DumitrascuEtAl2018} introduce an approximate variant of TS which uses P\'{o}lya-Gamma (PG) augmentation to admit efficient sampling. We adapt this to give an approximate TS policy for LCAT, PG-TS, in Algorithm \ref{alg::PGTS}. It utilises a Gibbs sampler for the unknown parameters, possible due to a parameter augmentation approach, and highly efficient rejection sampler for the augmenting parameters due to \cite{PolsonEtAl2013,WindleEtAl2014}. A full description of the Gibbs sampler is provided in Appendix \ref{app:gibbs}. 

In our algorithms we let the function \texttt{GIBBS}$(\mathbf{b},\mathbf{B},M,\mathcal{D},\theta)$ denote the use of said Gibbs sampler initialised at parameter $\theta$ to draw $M$ samples from the approximation of the posterior implied by a prior, $MVN_\Theta(\mathbf{b},\mathbf{B})$ which is a Gaussian with mean $\mathbf{b}$ and covariance $\mathbf{B}$, restricted to $\Theta$, and observed data $\mathcal{D}$.

\begin{algorithm}[htbp]
    \caption{PG-TS}
    \label{alg::PGTS}
    \hrule
    \vspace{0.2cm}    
        \textbf{Inputs:} Prior mean vector $\mathbf{b}$, Prior covariance matrix $\mathbf{B}$, Number of Gibbs iterations $M$.\\
        \vspace{0.2cm} \hrule \vspace{0.2cm}
        \textbf{Initialise:} $\mathcal{D}=\emptyset$, and $\theta^{(M)}_0\sim MVN_\Theta(\mathbf{b},\mathbf{B})$.\\
        
        \textbf{for} $t=1,2,\dots$ \textbf{do} \\
         \qquad  $\{\theta_t^{(1)},\dots,\theta_t^{(M)}\} \leftarrow$ \texttt{GIBBS}$(\mathbf{b},\mathbf{B},M,\mathcal{D},\theta_{t-1}^{(M)})$ \\  
         \qquad Receive context ${x}_t\in \mathbb{R}^d$ \\ 
         \qquad Select action $A_t = \alpha_t(\theta^{(M)}_t)$ \\
         \qquad \textbf{if} $A_t =1$ \textbf{do} \\ 
         \qquad \qquad Observe $\Phi_t(1)\in\{l_{01},l_{11}\}$ \\ 
         \qquad \qquad Augment $\mathcal{D} \leftarrow \mathcal{D} \cup \{ x_t,\Phi_t(1)\}$ \\ 
         \qquad \textbf{end if} \\ 
         \textbf{end for}
         \vspace{0.2cm}
      \hrule
      \vspace{0.2cm}
\end{algorithm}

\subsection{On the Impact of Approximate Inference} \label{sec::approx}

The regret results of the Section \ref{sec::theory} are based on exact sampling from the posterior. The PG-TS algorithm necessarily samples from an approximation of the posterior, to maintain a reasonable computational overhead. Recent work of \cite{PhanEtAl2019} has identified conditions under which sampling from an approximate posterior can lead to linear regret in multi-armed bandit problems. On the other hand, \cite{MayEtAl2012} have shown that sublinear regret in contextual bandits can be achieved without drawing samples from an \emph{exact} posterior - in fact that it suffices to sample from a distribution that converges around the true parameters in the limit. 

In this section we address the possible concern around use of an approximate sampler, by demonstrating that PG-TS does not meet the sufficient conditions identified by \cite{PhanEtAl2019}, and further adapt the results of \cite{MayEtAl2012} to LCAT to show that PG-TS obtains an asymptotically sublinear regret.

\cite{PhanEtAl2019} characterise approximate TS policies in terms of their $\upalpha$-divergence\footnote{Here we use the script $\upalpha$ to avoid confusion with $\alpha_t$, the optimal action selection function.}, defined for a pair of distributions $P$, $Q$ with densities $p(x)$, $q(x)$, and a coefficient $\upalpha\in\mathbb{R}$ as, \begin{equation}
    D_\upalpha(P,Q)=\frac{1-\int p(x)^\upalpha (1-q(x))^{1-\upalpha}\mathrm{d}x}{\upalpha(1-\upalpha)}.
\end{equation} The $\upalpha$-divergence generalises a number of divergences including $KL(Q,P)$ (when $\upalpha\rightarrow 0$) and $KL(P,Q)$ (when $\upalpha\rightarrow 1$), and can be related to the Total Variation (TV) distance via Pinsker's inequality. 

At a high level, the contribution of \cite{PhanEtAl2019} is to show that there exist approximate distributions $Q_t$ which satisfy $D_\upalpha(\Pi_t,Q_t)<\epsilon$ for a true posteriors $\Pi_t$ and constant $\epsilon>0$, but sampling from $Q_t$ at every time step results in linear regret. In effect this shows that $D_\upalpha(\Pi_t,Q_t)<\epsilon$ is not a sufficient condition for approximate TS according to $Q_t$ to inherit sublinear regret guarantees of exact TS according to $\Pi_t$.

In our setting then, let $\theta_t$ be the sample used in round $t$ under exact TS, and $\theta^{(M)}_t$ be the sample used in round $t$ under PG-TS. Let $\pi_t$, and $\pi_t^{(M)}$ be the densities associated with these random variables, assuming the same truncated multivariate Gaussian prior in both cases. Our first theoretical result of this section, Theorem \ref{thm::no_divergence} below, will demonstrate that in the limit (with respect to $t$) the $\upalpha$-divergence between $\pi_t$ and $\pi_t^{(M)}$ goes to $0$. Therefore the PG-TS scheme is not unreliable in the sense identified by \cite{PhanEtAl2019}.

Our second result, Theorem \ref{thm::maylike} below, uses the results of \cite{MayEtAl2012} to show that the expected regret of PG-TS is indeed asymptotically sublinear in $T$. Theorem 1 of \cite{MayEtAl2012} establishes sufficient conditions for asymptotic consistency of a randomised contextual bandit algorithm. Specifically, they consider the contextual bandit problem where context $x_t$ in a closed $\mathcal{X}$ is observed, and action $a_t$ in a finite set $\mathcal{A}$ is selected, inducing reward observation $r_t=f_{a_t}(x_t)+z_{t,a_t}$, where $f_a: \mathcal{X}\rightarrow \mathbb{R}$ are unknown continuous functions, and $z_{t,a}$ are zero-mean random variables. \cite{MayEtAl2012} establish conditions under which the sequence of chosen actions $\{a_1,a_2,\dots\}$ satisfies the following convergence criterion of \cite{yang2002randomized}, \begin{equation}
    \frac{\sum_{t=1}^T f_{a_t}(x_t)}{\sum_{t=1}^T f^*(x_t)}\rightarrow 1 \text{ a.s.}, \text{ as } t\rightarrow \infty, \label{eq::cc}
\end{equation} where $f^*(x)=\argmax_{a\in\mathcal{A}}f_a(x)$ is the optimal expected reward.

Both theorems make use of additional assumptions on the parameter space and context distribution, which require the following additional notation. For $\theta \in \mathbb{R}^d$ define the sets $\mathcal{X}_1(\theta)=\{x\in\mathcal{X}: \alpha(x,\theta)=1\}$, and $\mathcal{X}_0(\theta)=\mathcal{X}\setminus \mathcal{X}_1(\theta)$, of context vectors such that actions $1$ and $0$, respectively, are optimal. 

The following assumptions on the properties of $\mathcal{X}_1$ and $\mathcal{X}_0$ ensure that our approximate posterior distributions converge appropriately. Assumption \ref{assum:1} ensures repeated sampling, regardless of where the posterior mass initially gathers, and Assumption \ref{assum:2} ensures that some items of class $0$ will be observed among the revealed true labels.

\begin{assumption} \label{assum:1} Contexts are drawn i.i.d. from distribution $p_X$ on $\mathcal{X}$, and there exists $\delta>0$ such that $p_X(\mathcal{X}_1(\theta)),$ $p_X(\mathcal{X}_0(\theta))>\delta$ for every $\theta\in\Theta$.  \end{assumption}

\begin{assumption} \label{assum:2} For every $\theta\in\Theta$, there exists $x\in\mathcal{X}_1(\theta)$ such that $\sigma(x^\top\theta^*)<1$.
\end{assumption}

We are now ready to present our theoretical results relating to the performance of PG-TS. The proofs of the theorems are given in Appendices \ref{app::no_divergence} and \ref{app::maylike} respectively.

\begin{theorem} \label{thm::no_divergence}
Under Assumptions \ref{assum:1} and \ref{assum:2}, and for the densities $\pi_t$ and $\pi_t^{(M)}$ as defined above, we have \begin{equation}
    \lim_{T\rightarrow \infty} D_{KL}\left(\pi_T,\pi_T^{(M)}\right) =0.
\end{equation}
\end{theorem}

\begin{theorem}
Under Assumptions \ref{assum:1}, and \ref{assum:2}, the sequence of actions $\{A_t^{(M)}\}_{t=1}^\infty$ selected by PG-TS satisfies the convergence criterion \eqref{eq::cc}, and thus the regret of PG-TS is asymptotically sublinear. \label{thm::maylike}
\end{theorem}

To go further than these results - i.e. to establish that the regret  guarantees associated with exact TS carry to PG-TS - is likely to be much more complex. The most advanced results on the regret of approximate TS in simple multi-armed bandits, for instance, rely on complex Bayesian non-parametric theory, which, to the best of our knowledge, does not yet have a known analog applicable to contextual problems \citep{mazumdar2020thompson}.

\section{Information Directed Sampling} \label{sec::altapp}

The PG-augmentation scheme can also be used to devise an approximate Information Directed Sampling (IDS) scheme, based on the framework proposed by \cite{RussoVanRoy2018}. IDS algorithms are randomised policies which construct an action sampling distribution, in each round $t$, based on a trade-off of regret and information gain. They have been shown to enjoy a similar, or improved, theoretical and empirical performance to TS as well as a potential for generalisation to a wider range of partial monitoring problems, since they do not restrict themselves to selecting actions which have a non-zero probability of being optimal. \REV{It is likely that some similar regret guarantees to those derived for TS in Section \ref{sec::mainproof} could be derived for IDS, since it explicitly optimises the same quantity (information ratio) that is central in the analysis of TS. The analysis for IDS however needs to consider posterior expectations, rather than posterior samples, and is likely to be of a somewhat different flavour - and is beyond the scope of this paper.} 

Two general methods to select the IDS action sampling distribution have been proposed. Both are designed to trade-off between achieving a low expected regret, and a high expected information gain. We shall explain these in a more general bandit-type loss-minimisation setting where $\mathcal{A}_t\subseteq\mathcal{A}\subset\mathbb{R}^d$ denotes a (potentially continuous) action set at time $t$, $l:\mathcal{A}\rightarrow \mathbb{R}$ denotes an expected loss function, and  $\Delta_t:\mathcal{A}\rightarrow\mathbb{R}$ and $I_t:\mathcal{A}\rightarrow \mathbb{R}_{>0}$ compute the expected regret and expected information gain of actions with respect to the posterior distribution in round $t$.

The first variant of IDS, which is the main focus of \cite{RussoVanRoy2014, RussoVanRoy2018} and \cite{KirschnerEtAl2020}, chooses its action sampling distribution $\tilde\pi_t^{IDS}$ to satisfy, \begin{equation}
    \tilde\pi_t^{IDS} \in \argmin_{\pi \in \mathcal{D}(\mathcal{A}_t)} \tilde\Psi_t^{IDS}(\pi), \text{ where } \tilde\Psi_t^{IDS}(\pi) = \frac{\tilde\Delta_t(\pi)^2}{\tilde{I}_t(\pi)}. \label{eq::IDSdef}
\end{equation}
Here, $\mathcal{D}(\mathcal{A}_t)$ is a family of distributions over $\mathcal{A}_t$, and $\tilde\Delta_t$ and $\tilde{I}_t$ are analogs of the regret and information gain for distributions. Specifically, $\tilde\Delta_t(\pi)=\int_{a\in\mathcal{A}}\Delta_t(a)\mathrm{d}\pi(a)$, and $\tilde{I}_t(\pi)=\int_{a\in\mathcal{A}}I_t(a)\mathrm{d}\pi(a)$.

The second variant introduces a further tunable parameter $\lambda>0$ and characterises the trade-off by a difference rather than a ratio, selecting its action sampling distribution $\pi_t^{IDS}$ as follows, \begin{equation}
    \pi_t^{IDS} \in \argmin_{\pi \in \mathcal{D}(\mathcal{A}_t)} \Psi_t^{IDS}(\pi), \text{ where } \Psi_t^{IDS}(\pi) = {\Delta_t(\pi)^2}-\lambda{I_t(\pi)}. \label{eq::tuneIDSdef}
\end{equation} This second approach is mentioned in \cite{RussoVanRoy2014,RussoVanRoy2018}, but has received less attention elsewhere. It can be shown to inherit the theoretical properties of the first variant if $\lambda\geq \tilde\Psi_t^{IDS}(\tilde\pi_t^{IDS})$ for every $t\in[T]$. We will show that this second approach is, conceptually, better suited to our problem. Since the first variant (i.e. the policy using \eqref{eq::IDSdef}) has been more widely used we will refer to it as \emph{traditional} IDS, and the second (i.e. the policy using \eqref{eq::tuneIDSdef}) as \emph{tunable} IDS.

Notice that in the apple tasting problem, as there are only two actions in any round and as $A_t=0$ has no associated information gain, we have $I_t(\pi_p)=pI_t(1)$, for any Bernoulli action selection distribution $\pi_p$ where the probability of choosing class 1 is $p$. The optimisation problem \eqref{eq::IDSdef} can be reduced to a line search over $p \in(0,1)$, and the gradient of the objective can be written, \begin{equation*}
    \frac{\mathrm{d}\tilde\Psi_t^{IDS}(\pi_p)}{\mathrm{d}p}=\frac{p^2{I}_t(1)\left(\Delta_t(1)-\Delta_t(0)\right)^2-I_t(1)\Delta_t(0)^2}{I_t(1)p} \REV{=\frac{p^2\left(\Delta_t(1)-\Delta_t(0)\right)^2-\Delta_t(0)^2}{p}.}
\end{equation*} It follows that $\tilde\pi_t^{IDS}=\pi_{\tilde{p}_t}$, where \begin{equation*}
    \tilde{p}_t=\min\left(1,\frac{\Delta_t(0)}{|\Delta_t(1)-\Delta_t(0)|}\right).
\end{equation*} Notice that $\tilde{p}_t$ is independent of $I_t(1)$ (so long as $I_t(1)>0$), and secondly that for all $\Delta_t(1)\leq 2\Delta_t(0)$ traditional IDS chooses the label 1 with probability 1, even if $\Delta_t(1)>\Delta_t(0)$. 

\begin{remark} \label{rem::IDS} We see that for traditional IDS, the \emph{magnitude} of the information gain in a particular round is immaterial. As there is an action with no information, IDS prefers the information gaining action unless there is strong regret based reason to choose the no information action. Observe that $\Delta_t(1)$ must be three times as large as $\Delta_t(0)$ before traditional IDS would begin to prefer action 0. We argue that this property is not desirable in a contextual (or otherwise non-stationary) setting, where the information gain may change from round to round, and note that it occurs in more general settings.

Considering the more general form of $\tilde\Psi_t^{IDS}$ in \eqref{eq::IDSdef}, it is clear that $\tilde\pi^{IDS}_t$ is invariant to any uniform scaling of the expected information function $\tilde{I}_t$. This is a potentially hazardous property in a range of situations, but seems to have the most pronounced effect in the setting where (potentially optimal) zero-information actions exist.
\end{remark}

\REV{Somewhere around here, it would be useful to address the concern of Reviewer 3 and consider whether IDS (with the ability to compute posterior expectations) inherits the regret bound of TS.}

In what follows we consider tunable IDS. Here the action selection distribution does depend on the magnitude of the information in a particular round, making it more suitable for our contextual problem. In particular, we have $\pi_t^{IDS}=\pi_{p_t}$, where \begin{equation}
    p_t=\max\left(0,~\min\left(1,~\frac{\lambda I_t(1)}{2(\Delta_t(1)-\Delta_t(0))^2}-\frac{\Delta_t(0)}{\Delta_t(1)-\Delta_t(0)}\right)\right). \label{eq::TuneIDSParam}
\end{equation} 

\subsection{Polya-Gamma Information Directed Sampling}
As an alternative to our TS approach, we explore a tunable PG-IDS scheme, summarised in Algorithm \ref{alg::PGIDS}. 

The non-trivial difference between the PG-IDS and PG-TS schemes, from an implementation perspective, is the requirement to estimate posterior expectations rather than just draw a single sample from an approximate posterior. An exact IDS scheme would compute $\Delta_t(0)$, $\Delta_t(1)$, and $I_t(1)$ in each round according to, \begin{align*}
    \Delta_t(a) &= \int_{\theta\in\mathbb{R}^d} \left[\mu(a,\theta)-\min_{a'\in\{0,1\}}\mu(a',\theta)\right] \mathrm{d}\pi_{t-1}(\theta), \quad a \in \{0,1\},
\end{align*} and \begin{align*}
I_t(1) &= \mathbb{P}_{t-1}\left(C_t=1\right)KL(\pi_{t-1},\pi_{t-1}|C_t=1) + \mathbb{P}_{t-1}\left(C_t=0\right)KL(\pi_{t-1},\pi_{t-1}|C_t=0).
\end{align*} Such computations are of course infeasible due to the intractability of the posterior discussed in Section \ref{sec::PGTS}. 

Instead, we rely on Monte Carlo estimators of these quantities. For expected regrets we have \begin{align}
    \bar\Delta_t(a) &= \frac{1}{M}\sum_{m=1}^M\left[\mu(a,\theta^{(m)})-\min_{a'\in\{0,1\}}\mu(a',\theta^{m})\right], \quad a \in \{0,1\}. \label{eq::RegEst}
\end{align} Before defining the information gain, introduce the notation, $$\ell(x,\theta,c)=\left(\sigma(x^\top\theta)\right)^{c}\left(1-\sigma(x^\top\theta)\right)^{1-c}$$ as the likelihood contribution of the pair $x,c$ for a given $\theta$. To estimate $I_t(1)$ we require approximations of the normalising constants of the current posterior $\pi_{t-1}$, and the potential posteriors given the two possible updates. Define, \begin{align*}
    \bar{D}_t &= \frac{1}{M}\sum_{m=1}^M\left[\prod_{s\in[t-1]:A_s=1} \ell(x_s,\theta^{(m)},C_s)\right], \text{ and }\\
    \bar{D}_{t,c} &= \frac{1}{M}\sum_{m=1}^M\left[\ell(x_t,\theta^{(m)},c)\prod_{s\in[t-1]:A_s=1} \ell(x_s,\theta^{(m)},C_s)\right],
\end{align*} for $c\in\{0,1\}$. These are used to estimate the KL divergences between the current posterior and the two potential posteriors in the next round. Define said estimates as, \begin{align*}
    \bar{K}_{t,c} = \frac{1}{M}\sum_{m=1}^M \log\left(\frac{\bar{D}_{t,c}}{\bar{D}_{t-1}\ell(x_t,\theta^{(m)},c)} \right), \quad c\in\{0,1\}.
\end{align*} Finally our estimate of the expected information gain follows as,
\begin{align}
    \bar{I}_t(1) &= \bar{K}_{t,1}\sum_{m=1}^{M}\frac{\ell(x_t,\theta^{(m)},1)}{M} + \bar{K}_{t,0}\sum_{m=1}^{M}\frac{\ell(x_t,\theta^{(m)},0)}{M}. \label{eq::InfEst}
\end{align}  We investigate the performance of PG-IDS alongside PG-TS in the next section.

\begin{remark}
Both the PG-TS and PG-IDS approaches given in Algorithm \ref{alg::PGTS} and \ref{alg::PGIDS} are perhaps the most straightforward possible in terms of their use of the Gibbs samples. It may be beneficial for instance to allow $M$ to vary as a function of $t$, to remove some burn-in from the sample $\{\theta_t^{(1)},\dots,\theta_t^{(M)}\}$ before computing posterior expectations, in the case of PG-IDS, or to use separate samples to compute the regret estimates, and information gain. In the following section we have found strong empirical performance is achieved without such modifications, but it may be an interesting direction for future research to investigate whether these play a material role in the algorithms' performance. 
\end{remark}

\begin{algorithm}[htbp]
    \caption{PG-IDS}
    \label{alg::PGIDS}
    \hrule
    \vspace{0.2cm}  
        \textbf{Inputs:} Prior mean vector $\mathbf{b}$, Prior covariance matrix $\mathbf{B}$, Number of Gibbs iterations $M$, IDS-tuning parameter  $\lambda$. \\
            \vspace{0.2cm} \hrule \vspace{0.2cm}
        \textbf{Initialise:} $\mathcal{D}=\emptyset$, and $\theta_0^{(M)}\sim MVN_\Theta(\mathbf{b},\mathbf{B})$. \\
        
        \textbf{for} $t=1,2,\dots$ \textbf{do} \\
         \qquad $\{\theta_t^{(1)},\dots,\theta_t^{(M)}\} \leftarrow$ \texttt{GIBBS}$(\mathbf{b},\mathbf{B},M,\mathcal{D},\theta_{t-1}^{(M)})$  \\
         \qquad Receive context $\mathbf{x}_t\in \mathbb{R}^d$ \\ 
         \qquad Compute regret estimates $\bar\Delta_t(0)$ and $\bar\Delta_t(1)$ according to \eqref{eq::RegEst} \\
         \qquad Compute information gain estimate $\bar{I}_t(1)$ according to \eqref{eq::InfEst} \\
         \qquad Compute IDS parameter $p_t$ according to \eqref{eq::TuneIDSParam}  using $\bar\Delta_t(0),\bar\Delta_t(1),\bar{I}_t(1)$\\ 
         \qquad Select action $a_t \sim Bern(p_t)$ \\ 
         \qquad \textbf{if} $a_t =1$ \textbf{do} \\ 
         \qquad \qquad Observe $l_t \in \{l_{01},l_{11}\}$ \\ 
         \qquad \qquad Augment $\mathcal{D} \leftarrow \mathcal{D} \cup \{x_t,\Phi_t(1)\}$ \\ 
         \qquad \textbf{end if} \\ 
         \textbf{end for}
         \vspace{0.2cm}
      \hrule
      \vspace{0.2cm}
\end{algorithm}

\section{Simulations} \label{sec::experiments}
We compare the PG-TS scheme in Algorithm \ref{alg::PGTS} with a number of other algorithms. Firstly, the PG-IDS scheme given in Algorithm \ref{alg::PGIDS} and second, an $\epsilon$-Greedy algorithm. The $\epsilon$-Greedy approach chooses the action optimal with respect to the maximum likelihood estimate with probability $1-\epsilon$, otherwise it chooses randomly with equal probability. Finally, we consider the {CBP-SIDE} algorithm of \cite{BartokSzepesvari2012}, as investigated empirically in \cite{lienert2013exploiting}. Pseudocode for the adaptation of this approach to the apple tasting problem is given in Appendix \ref{app:CBP}. Fast implementation of the PG-based algorithms is possible thanks to the `BayesLogit' package in R \citep{polson2019package}. 

For \REV{tunable} PG-IDS, we have selected the tunable parameter $\lambda$ via a further empirical comparison outlined in Appendix \ref{app::parameter}. The choices $\lambda=0.05$, is argued to give the most robust performance among a range of possible values across various instances of our problem. \REV{For $\epsilon$-Greedy we use $\epsilon=0.1$.}

We consider three examples in terms of the true parameter, and context vector distribution, summarised below. \begin{enumerate}
    \item[(i)] Each dimension of the parameter $\theta^*$ is sampled uniformly from the interval $[-1,1]$, with $d=5$. Contexts are drawn from a zero-mean multivariate Gaussian with identity covariance matrix. We choose $l_{11}=0.05$, $l_{01}=0.4$, $l_{10}=1$, and $T=500$. 
    \item[(ii)] Each dimension of the parameter $\theta^*$ is sampled uniformly from the interval $[-1,1]$ with probability 0.75, or fixed to 0 with probability 0.25. The number of dimensions is $d=20$, and again contexts are drawn as in problem (i) but with common context variance of 8. We choose $l_{11}=0.1$, $l_{01}=0.7$, $l_{10}=1$, and $T=1000$.
    \item[(iii)] $\theta^*=1$ and contexts are sampled from a Gaussian distribution with  standard deviation $0.025$. The mean linearly increases from $\mu=-0.1$ to $\mu=0$ over the $T=500$ rounds. We choose $l_{11}=0$ and $l_{01}=l_{10}=1$.
\end{enumerate} Problems (i) and (ii) represent typical apple tasting scenarios, and allow the comparison of our various algorithms. Problem (iii) gives a simple, uncluttered example of a scenario where the traditional variant of IDS performs poorly.

\begin{figure}
    \centering
    \begin{subfigure}{\textwidth}
    \centering
     \includegraphics[width=0.8\textwidth]{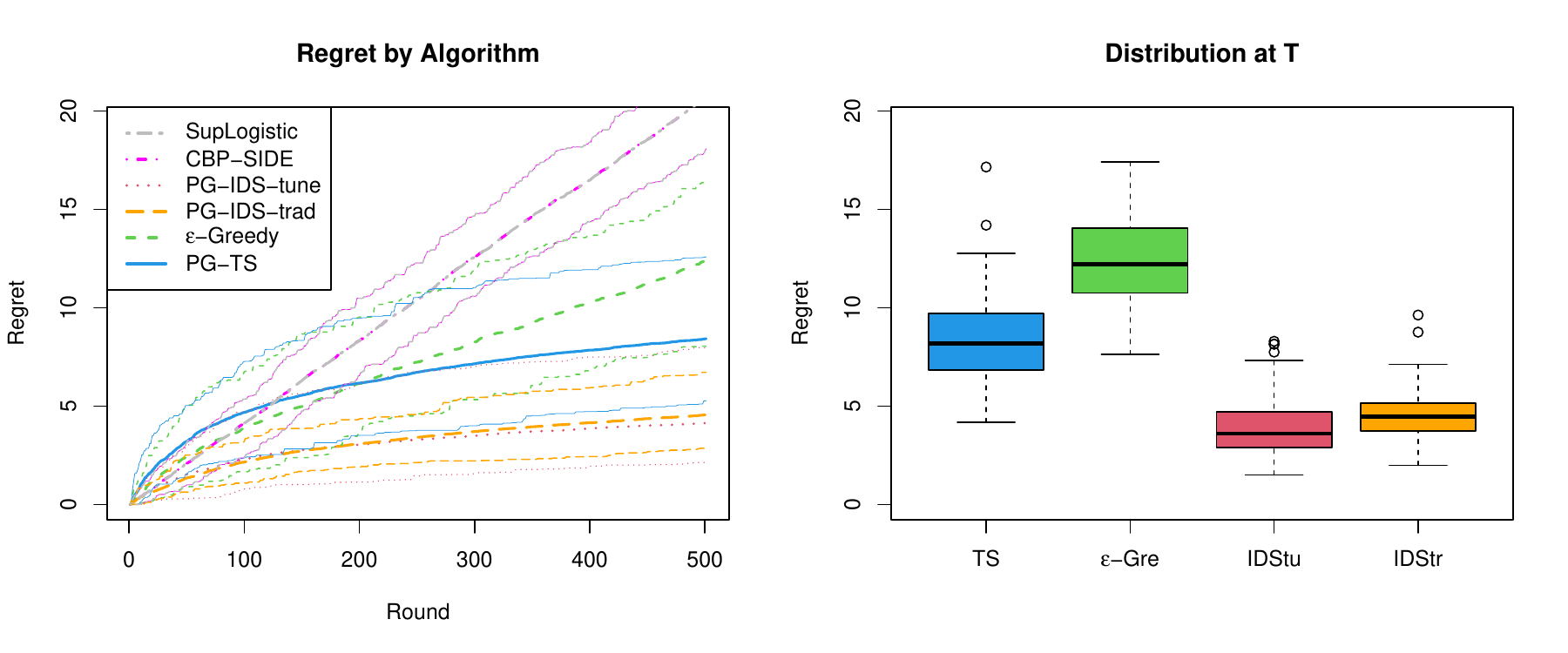}
     \caption{Problem (i)}
    \end{subfigure}
    \begin{subfigure}{\textwidth}
    \centering
     \includegraphics[width=0.8\textwidth]{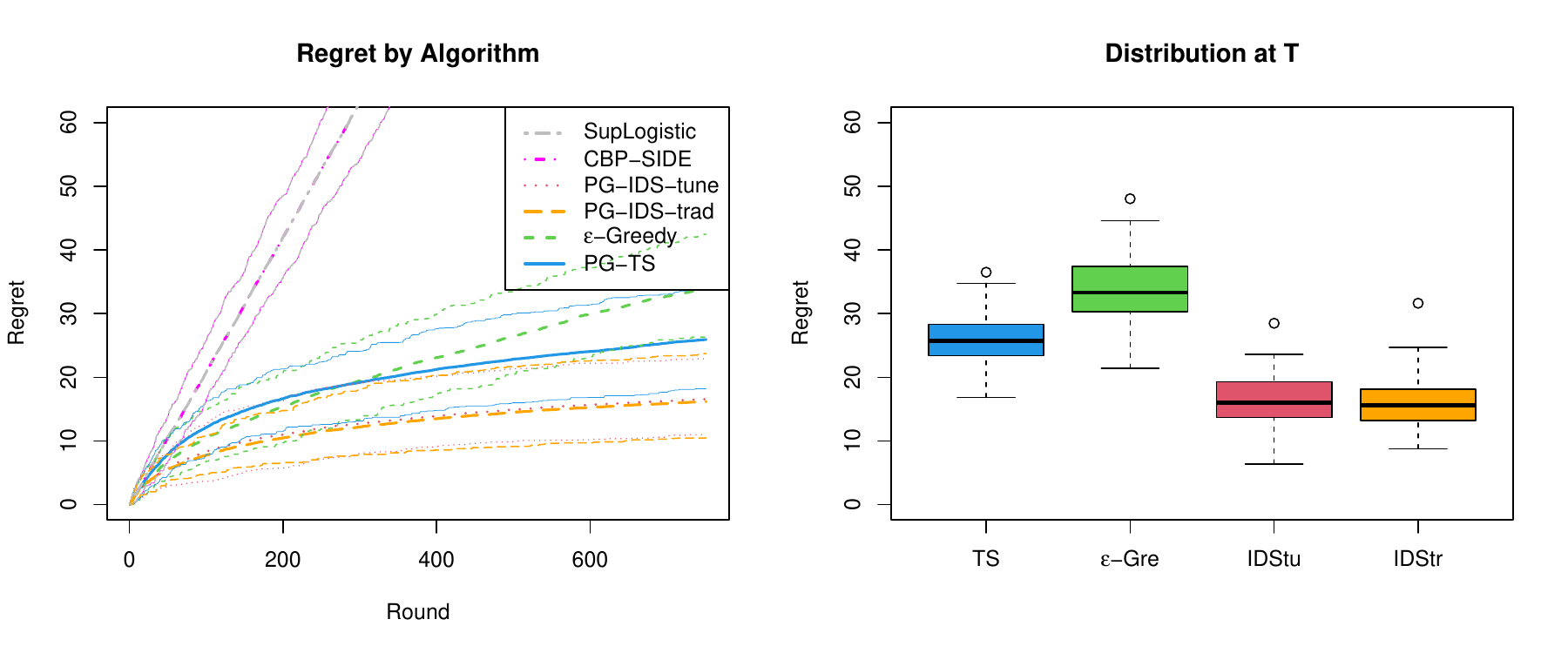}
     \caption{Problem (ii)}
    \end{subfigure}
    \begin{subfigure}{\textwidth}
    \centering
     \includegraphics[width=0.8\textwidth]{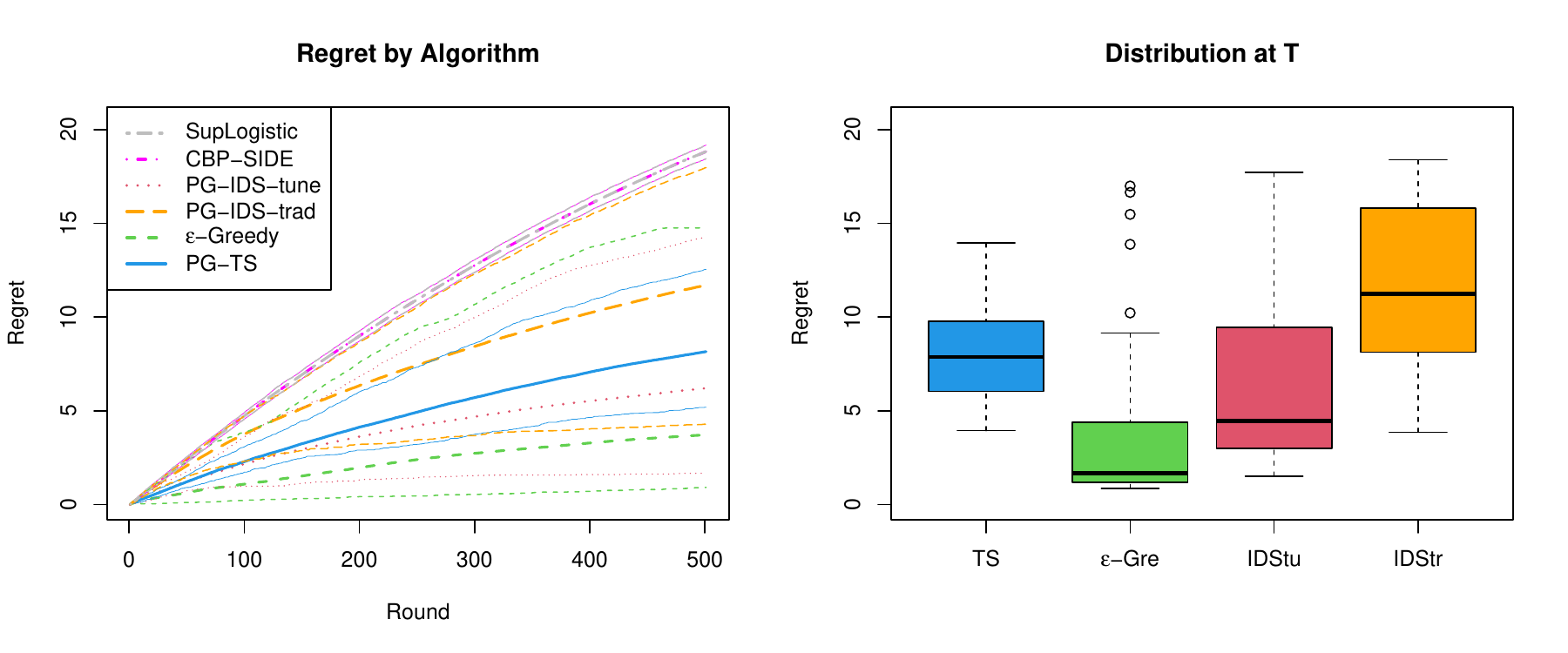}
     \caption{Problem (iii)}
    \end{subfigure}
    \caption{Regret of algorithms on Problems (i), (ii) and (iii), over 50 replications. The green lines denote the $\epsilon$-Greedy policy with $\epsilon=0.1$, the yellow lines denote the traditional PG-IDS policy, the red lines denote the tunable PG-IDS policy with $\lambda=0.2$, the magenta lines denote the CBP-SIDE policy, \REV{the gray lines denote the SupLogistic policy}, and the blue lines denote the PG-TS policy. In each case 90\% empirical confidence regions are plotted around the median trajectory. The boxplots in the right-hand panel show the distribution of the final regret at time $T$ \REV{for the four most successful algorithms}.}
    \label{fig:probi}
\end{figure}

\begin{figure}
    \centering
    \begin{subfigure}[b]{0.45\textwidth}
    \includegraphics[width=\textwidth]{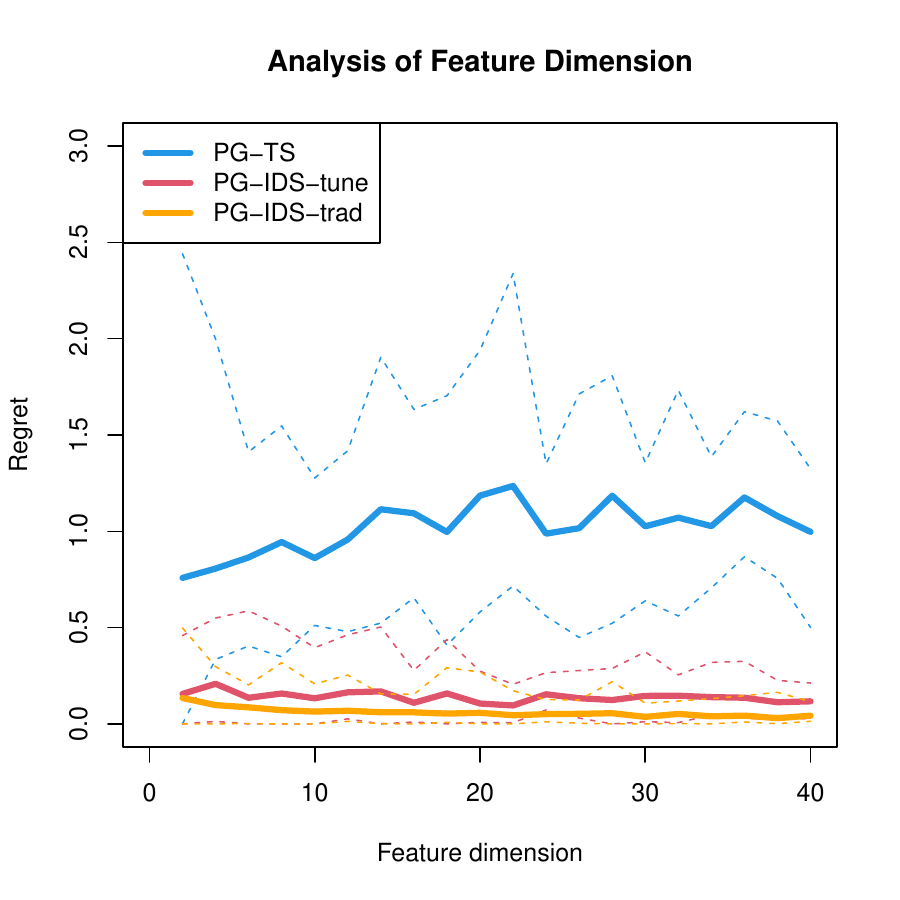}
    \caption{Regret of algorithms scaled by $\sqrt{d}$, plotted as $d$ varies.}
    \end{subfigure}
    \hfill
    \begin{subfigure}[b]{0.45\textwidth}
    \includegraphics[width=\textwidth]{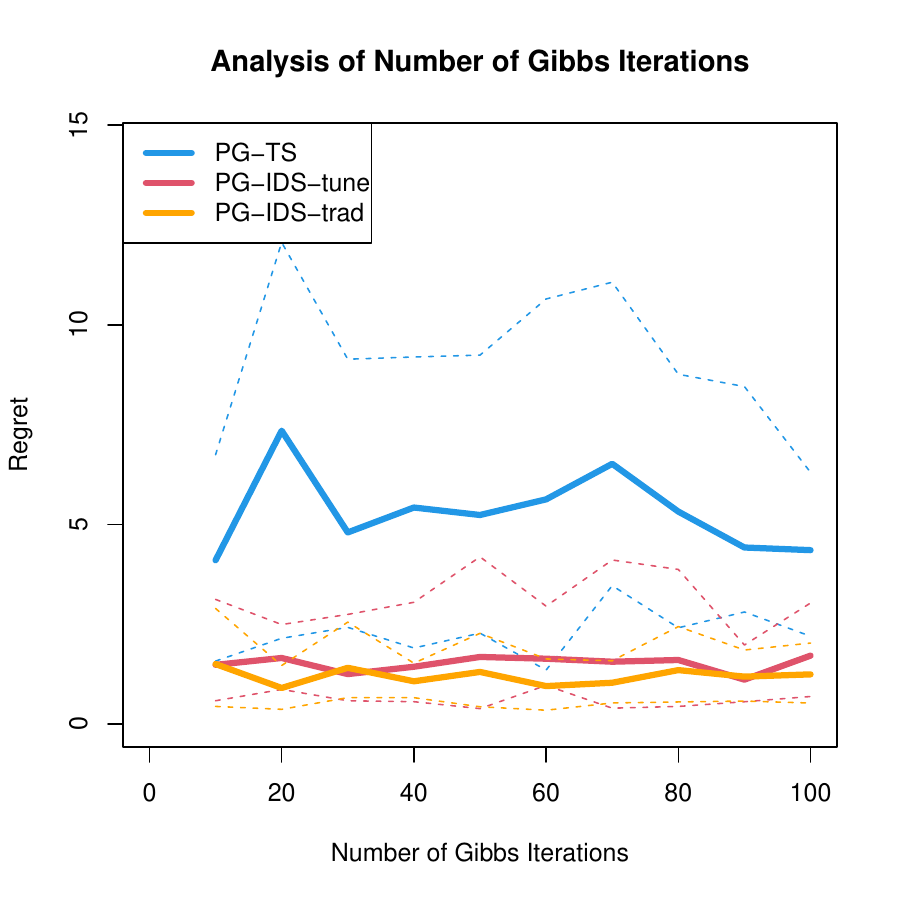}
    \caption{Regret of algorithms plotted as $M$ varies.\\}
    \end{subfigure}
    \caption{Plots showing the effects of problem and algorithm parameters on regret.}
    \label{fig:dimension}
\end{figure}

Figure \ref{fig:probi} shows the performance of the algorithms on the problems described above. For problems (i) and (ii) we see similar behaviour, the two PG-IDS algorithms perform similarly and incur a lower average regret than the other approaches. PG-TS is successful in learning to select optimal actions but does so more slowly, and \REV{SupLogistic}, CBP-SIDE and $\epsilon$-Greedy incur a regret which appears to grow linearly throughout the experiment. \REV{The } $\epsilon$-Greedy \REV{algorithm} could perhaps be improved by allowing $\epsilon$ to decay as a function of $t$, but this represents yet another possibly problem-dependent tuning step. As for \REV{SupLogistic and} CBP-SIDE, the observed behaviour is a result of \REV{their} overly conservative approaches. This is confirmed below by considering the precision and recall of the algorithms. \REV{Their regret curves are observed to be  very close, or indeed overlapping, because of this near-deterministic, conservative behaviour.}

In problem (iii) where every context is sampled close to the classification boundary, with a strong tendency towards class 0, a different behaviour is observed. Besides PG-TS, \REV{SupLogistic}, and CBP-SIDE, each of the algorithms displays a more variable performance, and the traditional variant of PG-IDS incurs noticeably larger regret than PG-TS and the tunable variant of PG-IDS. The mean regret of PG-TS and the tunable variant of PG-IDS are very similar. While $\epsilon$-Greedy sometimes outperforms the more principled approaches, due to fixing an accurate initial estimate of $\theta^*$, this overly exploitative behaviour sometimes fails and leads to highly skewed distribution of regret. 

The phenomenon of a greedy approach sometimes outperforming more complex attempts to balance exploration and exploitation in contextual problems is not unprecedented. Indeed, in the contextual bandit literature, a number of recent works \citep{BastaniEtAl2017,KannanEtAl2018,RaghavanEtAl2020,jedor2021greedy} have observed that if the contexts arising are sufficiently variable, a greedy decision-maker can gain appropriate information without explicitly attempting to explore in favour of exploiting.

The regret is presumed to be the true measure of interest in the apple tasting problem, and to appropriately weight the impact of false positives and false negatives. Nevertheless it is informative to see in what proportion the algorithms make the two classes of error. To this end, in Tables \ref{tab::1}, \ref{tab::2}, and \ref{tab::3} we report the \emph{precision} and \emph{recall} of the apple tasting algorithms. The precision is the proportion of true class 1 examples among all those labelled class 1 by the algorithm, the recall is the proportion of class 1 examples correctly labelled as class 1 by the algorithm.

We observe that CBP-SIDE \REV{and SupLogistic} always have a recall of $1.00$, and while we would expect this to reduce in problems with a longer horizon, this is indicative of the fact that CBP-SIDE's confidence region for $\theta^*$ is so wide, \REV{and SupLogistic's required data collection is so large,} that they use action 1 in almost every instance. As such they correctly labels all true class 1 instances, but misclassify a large (relative to the other approaches) proportion of class 0 instances as being class 1.

In problems (i) and (ii) most of the other approaches have well balanced precision and recall, suggesting that misclassifications of both types occur with roughly similar frequency. In problem (iii), however, the traditional variant of PG-IDS shows a similar behaviour to CBP-SIDE, in that its recall is much larger than the algorithms with smaller average regret. This is likely because it suffers the issue postulated in Section \ref{sec::altapp}. Specifically, that because of the construction of the traditional IDS sampling distribution, it continues to favour the information gaining action even when the regret of the no-information action is smaller, and the information gained per observation is minimal.

\begin{table}[ht]
\parbox{0.49\linewidth}{
\centering
\begin{tabular}{rrr}
  \hline
 Algorithm & Precision & Recall \\ 
  \hline
PG-TS & 0.93 & 0.90 \\ 
  PG-IDS-tune & 0.94 & 0.94 \\ 
  PG-IDS-trad & 0.91 & 0.96 \\ 
  Greedy & 0.93 & 0.91 \\ 
  CBP-SIDE & 0.74 & 1.00 \\ 
  \REV{SupLogistic} & \REV{0.74} & \REV{1.00} \\
   \hline
\end{tabular}
\caption{Precision and Recall on Problem (i)}
\label{tab::1}
}
\hfill
\parbox{0.49\linewidth}{
\centering
\begin{tabular}{rrr}
  \hline
 Algorithm & Precision & Recall \\ 
  \hline
PG-TS & 0.89 & 0.88 \\ 
  PG-IDS-tune & 0.91 & 0.92 \\ 
  PG-IDS-trad & 0.88 & 0.95 \\ 
  Greedy & 0.88 & 0.87 \\ 
  CBP-SIDE & 0.53 & 1.00 \\ 
  \REV{SupLogistic} & \REV{0.53} & \REV{1.00} \\
   \hline
\end{tabular}
\caption{Precision and Recall on Problem (ii)}
\label{tab::2}
} \\
\vspace{0.7cm}

\centering
\parbox{0.59\linewidth}{
\centering 
\begin{tabular}{rrr}
  \hline
 Algorithm & Precision & Recall \\ 
  \hline
PG-TS & 0.13 & 0.56 \\ 
  PG-IDS-tune & 0.26 & 0.75 \\ 
  PG-IDS-trad & 0.17 & 0.92 \\ 
  Greedy & 0.49 & 0.85 \\ 
  CBP-SIDE & 0.10 & 1.00 \\
  \REV{SupLogistic} & \REV{0.05} & \REV{1.00} \\
   \hline
\end{tabular}
\caption{Precision and Recall on Problem (iii)}
\label{tab::3}
}
\end{table}

Finally, we have also investigated the effect of the dimension $d$ of the unknown parameter, and of the number of Gibbs iterations $M$, on the regret of the algorithms. Figure \ref{fig:dimension} displays the results of these experiments. In the setting of Figure \ref{fig:dimension}(a), the dimension is varied between $d=2$ and $d=40$, with the elements of $\theta^*$ being drawn uniformly at random from the interval $[-1,1]$, \REV{contexts $x_t$ from a zero-mean multivariate Gaussian with covariance matrix\footnote{\REV{This choice ensures that the expectation and variance of $x_t\theta^*$ is constant as $d$ increases and the regret across problems can be fairly compared.}} $d^{-1/2}I_d$ (where $I_d$ is the $d$-dimensional identity matrix) and $M$ fixed at 15.} \REV{The setting of Figure \ref{fig:dimension}(b) is identical to that of Problem (i); we varied} the number of Gibbs iterations per time-step between $M=10$ and $M=100$. 

In Figure \ref{fig:dimension}(a) we plot the mean regret for each investigated dimension $d$ and algorithm with a 90\% empirical confidence interval, all scaled by $\sqrt{d}$ to investigate the relationship between regret and dimension. We see a mostly constant relationship, within the tolerance of statistical noise, which seems to validate the theoretical result on regret. In Figure \ref{fig:dimension}(b) we see again see a fairly constant relationship, suggesting that the performance of the algorithms is fairly robust to the number of Gibbs iterations. This is an encouraging finding as it suggests a more computationally burdensome choice of the parameter $M$ is not necessary for strong performance.

\section{Conclusion}

In this paper we have explored the use of heuristic Bayesian decision-making rules for logistic contextual apple tasting problems. We have shown that both Thompson Sampling and Information Directed Sampling methods are highly efficient for such problems, and indeed more so than confidence bound based approach of \cite{BartokSzepesvari2012}. We have also established a theoretical justification of the strong performance of Thompson Sampling through a bound on its Bayesian regret, and of P\'{o}lya-Gamma Thompson Sampling by considering its asymptotic behaviour. The extension of these results to Information Directed Sampling will be significantly more complex due to the choice of actions with respect to posterior expectations rather than individual samples\footnote{Note that this difficulty is specific to the present setting where $\Theta$ is not a finite set. The challenge arises in defining a suitable analog of the compressed random variable $\tilde\theta^*_t$ as in the proof of Theorem \ref{thm::TSBR}.}. 

\REV{We have shown that the existence of a non-informative action in apple tasting can inhibit the performance of traditional Information Directed Sampling, and it would be of interest to explore further more complex settings (in partial monitoring, reinforcement learning, etc.) where this effect could be even more pronounced.} Other further research may explore the application of Bayesian approaches to more complex contextual partial monitoring problems, incorporating the insights of \cite{LattimoreSzepesvari2019}, or to other variants of apple tasting, such as the fairness-enforcing variant considered by \cite{bechavod2019equal}, or batched setting of \cite{jiang2020learning}.

\textbf{Acknowledgements:} {The authors gratefully acknowledge the support of the Next Generation Converged Digital Infrastructure project (EP/R004935/1) funded by the EPSRC and BT, and thank Fabrizio Leisen, Christopher Nemeth, Ciara Pike-Burke, and Christopher Sherlock for helpful conversations during the preparation of this manuscript.}

\bibliographystyle{apalike}
\bibliography{PostdocRefs}

\appendix

\section{Proof of Proposition \ref{prop::thetatilde}} \label{app::thetatilde} \REV{The first step constructs the distribution of the random variable $\tilde\theta_t^*$, conditional on $\phi_\epsilon$. Subsequently, we show that properties of the conditional distribution carry to the unconditioned giving the stated results. 

For any round $t\in[T],$ its $x_t$, and partition cell $k \in [K]$, we define the subsets, \begin{align*}
    \Theta_{k,t}^1 = \left\{ \theta\in\Theta_k : \alpha_t(\theta)=1\right\}, \text{ and } ~\Theta_{k,t}^0 = \left\{ \theta\in\Theta_k : \alpha_t(\theta)=0\right\},
\end{align*} of parameters associated with selection of the informative and non-informative action respectively. Consider the case where both $\pi_{t-1}(\Theta_{k,t}^0) >0$, and $\pi_{t-1}(\Theta_{k,t}^1) >0$. Let $\theta_{k,t}^1\in\Theta_{k,t}^1$ and $\theta_{k,t}^0\in\Theta_{k,t}^0$ be arbitrary points chosen (necessarily) with knowledge of $x_t$. We define $\tilde\theta_t^*$ to have distribution, $\tilde\pi_{t-1}$ satisfying, \begin{align*}
\tilde\pi_{t-1}\left(\tilde\theta_t^*=\theta_{k,t}^1~|~\phi_\epsilon=k\right)&=\pi_{t-1}\left(\Theta_{k,t}^1~|~\phi_\epsilon=k\right), \\
\tilde\pi_{t-1}\left(\tilde\theta_t^*=\theta_{k,t}^0~|~\phi_\epsilon=k\right)&=1-\pi_{t-1}\left(\Theta_{k,t}^1~|~\phi_\epsilon=k\right).
\end{align*} In the case where either $\pi_{t-1}(\Theta_{k,t}^0)=0$ or $\pi_{t-1}(\Theta_{k,t}^1)=0$ we choose a single point $\theta_{k,t}$ in $\Theta_k$ arbitrarily which and have $\tilde\pi_{t-1}(\tilde\theta_t^*=\theta_{k,t}~|~\phi_\epsilon=k)=1$.

We have Property (i) as an immediate consequence of these construction, since although $\tilde\theta_t^*$ is defined with reference to the \emph{posterior density} on $\theta^*$, it is independent of its \emph{realisation}. Aggregating over cells of the partition, Property (iv) also holds by construction.

As a consequence of the logistic model, the sets $\Theta_{k,t}^1$ and $\Theta_{k,t}^0$ are separated by the hyperplane $x_t^\top\theta = \sigma^{-1}((1+l_{01}-l_{11})^{-1})$ and the functions $\mathbb{E}_{t-1}(\mu_t(\alpha_t(\theta),\theta^*)$ and $I_{t-1}(\phi_\epsilon;(\theta_t, \Phi_t(\alpha_t(\theta)))$ are constant in $\theta$ within each of $\Theta_{k,t}^0$ and $\Theta_{k,t}^1$. As such, we have equality of the expected losses, \begin{align}
    &\tilde\pi_{t-1}(\theta_{k,t}^1~|~\phi_\epsilon=k) \mathbb{E}_{t-1}(\mu_t(\alpha_t(\theta_{k,t}^1),\theta^*)) + (1-\tilde\pi_{t-1}(\theta_{k,t}^1~|~\phi_\epsilon=k)) \mathbb{E}_{t-1}(\mu_t(\alpha_t(\theta_{k,t}^0),\theta^*)) \nonumber \\
    &\quad \quad \quad = \mathbb{E}_{t-1}(\mu_t(\alpha_t(\theta_t),\theta^*)), \label{eq::equalexp}
\end{align} expected information gains, \begin{align} 
    &\tilde\pi_{t-1}(\theta_{k,t}^1~|~\phi_\epsilon=k) I_{t}(\phi_\epsilon;\Phi_t(\alpha_t(\theta_{k,t}^1))) + (1-\tilde\pi_{t-1}(\theta_{k,t}^1~|~\phi_\epsilon=k)) I_{t}(\phi_\epsilon;\Phi_t(\alpha_t(\theta_{k,t}^0))) \nonumber \\
    &\quad \quad \quad  = I_{t}(\phi_\epsilon;\Phi_t(\alpha_t(\theta_t))), \label{eq::equalinf} 
\end{align} and probability of selecting the informative action \begin{align} 
    &\tilde\pi_{t-1}(\theta_{k,t}^1~|~\phi_\epsilon=k) \mathbb{I}(\theta_{k,t}^1 \in \tilde\Theta_t) + (1-\tilde\pi_{t-1}(\theta_{k,t}^1~|~\phi_\epsilon=k)) \mathbb{I}(\theta_{k,t}^0 \in \tilde\Theta_t) \nonumber \\
    &\quad \quad \quad = \int_{\theta \in \Theta_k} \mathbb{I}(\theta \in \tilde\Theta_t)\mathrm{d}\pi_{t-1}(\theta_t=\theta ~|~ \theta^*\in \Theta_k). \nonumber
\end{align}

To show Property (ii), we first define, for every $k\in[K]$ and $t\in[T]$, \begin{align*}
    D_{k,t} &= r_{k,t}\mathbb{E}_{t-1}\left(\mu_t\left(\alpha_t\left(\theta_{k,t}^1\right),\theta^*\right)\right) + (1-r_{k,t})\mathbb{E}_{t-1}\left(\mu_t\left(\alpha_t\left(\theta_{k,t}^0\right),\theta^*\right)\right) \\
    &\quad \quad \quad \quad - \mathbb{E}_{t-1}\left(\mu_t\left(\alpha_t(\theta_t),\theta^*\right)~|~\theta_t\in\Theta_k\right).
\end{align*} By \eqref{eq::equalexp}, each $D_{k,t}= 0$. Thus, we also have, \begin{equation*}
    \mathbb{E}_{t-1}\left(\mu_t(\alpha_t(\tilde\theta_t),\theta^*)-\mu_t(\alpha(\theta_t),\theta^*)\right) = \sum_{k=1}^K P(\theta_t\in\Theta_k)D_{k,t} = 0.
\end{equation*} Property (ii) then follows using this result in the first inequality, \begin{align*}
    &~ \mathbb{E}_{t-1}\left(\mu_t\left(\alpha_t(\theta_t),\theta^*\right)-\mu_t\left(\alpha_t(\theta^*),\theta^*\right) \right) - \mathbb{E}_{t-1}\left(\mu_t\left(\alpha_t(\tilde\theta_t),\theta^*\right)-\mu_t\left(\alpha_t(\tilde\theta^*_t),\theta^*\right)\right) \\
    &=\mathbb{E}_{t-1}\left(\mu_t\left(\alpha_t(\tilde\theta^*_t),\theta^*\right)-\mu_t\left(\alpha_t(\theta^*),\theta^*\right) \right) - \mathbb{E}_{t-1}\left(\mu_t\left(\alpha_t(\tilde\theta_t),\theta^*\right)-\mu_t\left(\alpha_t(\theta_t),\theta^*\right)\right) \\
    &= \mathbb{E}_{t-1}\left(\mu_t\left(\alpha_t(\tilde\theta^*_t),\theta^*\right)-\mu_t\left(\alpha_t(\theta^*),\theta^*\right) \right) \leq \epsilon.
\end{align*} Here the inequality holds since $\tilde\theta_t^*$ and $\theta^*$ are always in the same cell of the partition, and (ii) follows by simple rearrangement.

Considering the information gain, we show Property (iii) as follows, where the second and final equalities use the independence  of $\theta_t$ and $\tilde\theta_t$ from $\phi_\epsilon$ conditioned on $\mathcal{H}_{t-1}$, and the fourth uses \eqref{eq::equalinf}, \begin{align*}
    I_{t}\left(\phi_\epsilon;\left(\tilde\theta_t,\Phi_t(\alpha_t(\tilde\theta_t)\right)\right) &=  I_{t}\left(\phi_\epsilon;\tilde\theta_t\right)+  I_{t}\left(\phi_\epsilon;\left(\Phi_t(\alpha_t(\tilde\theta_t)~|~\tilde\theta_t\right)\right) \\
    &= I_{t}\left(\phi_\epsilon;\left(\Phi_t(\alpha_t(\tilde\theta_t)~|~\tilde\theta_t\right)\right) \\
    &=\sum_{k=1}^K\sum_{i=1}^2 {P}\left(\theta_t\in\Theta_k\right){P}\left(\tilde\theta_t=\theta_i^{k,t}~|~\theta_t\in\Theta_k\right) I_{1}\left(\phi_\epsilon;\Phi_t(\alpha_t(\tilde\theta_i^{k,t})\right) \\
    &= \sum_{k=1}^K {P}\left(\theta_t\in\Theta_k\right) I_{t-1}\left(\phi_\epsilon;\Phi_t(\alpha_t(\theta_t))~|~\theta_t\in\Theta_k\right) \\
    &= {I}_{t}\left(\phi_\epsilon;\Phi_t(\alpha_t(\theta_t))\right) = I_{t}\left(\phi_\epsilon;\left(\theta_t,\Phi_t(\alpha_t(\theta_t)\right)\right). 
\end{align*}}

\section{Proof of Proposition \ref{prop::infratiobound}} \label{app::infratiobound} Recall the definition of the information ratio of the random variables $\tilde\theta_t^*$ and $ \tilde\theta_t$ as
\begin{equation*}
    \Gamma_t\left(\tilde\theta_t^*, \tilde\theta_t \right) = \frac{\left[\mathbb{E}_{t-1}\left( \mu_t\left(\alpha_t(\tilde\theta_t),\theta^*\right)-\mu_t\left(\alpha_t(\tilde\theta^*_t),\theta^*\right) \right)\right]^2}{\REV{I_{t}}\left(\tilde\theta_t^*; (\tilde\theta_t,\Phi_t(\alpha_t(\tilde\theta_t)))\right)},
\end{equation*}
where the expectation in the numerator is taken over the random variables $\tilde\theta_t^*, \tilde\theta_t$ and $\theta^*$ conditional on the history $\mathcal{H}_{t-1}$ (recall that $\tilde{\theta}^*_t$ is the compressed version of $\theta^*$, whereas $\tilde{\theta}_t$ has the same marginal distribution as $\tilde{\theta}^*_t$ but is conditionally independent of $\theta^*$ and $\tilde{\theta}^*_t$ given $\mathcal{H}_{t-1}$). It is worth noting that although the information gain in a particular round \emph{can} take value zero, the quantity in the denominator is an expectation of the mutual information over the distribution of $\alpha_t(\tilde\theta_t)$ and the signal $\Phi_t(\alpha_t(\tilde\theta_t))$ and as such will be non-zero so long as $\pi_{t-1}(\REV{\tilde\Theta_t})>0$.

To bound $\Gamma_t$, we first rewrite the root of the numerator in the information ratio. 
Denote by $\tilde{\pi}_t(\tilde{\theta}_t)$ and $\tilde{\pi}^*_t(\tilde{\theta}^*_t)$ the marginal probability mass functions of $\tilde{\theta}_t$ and $\tilde{\theta}^*_t$ respectively conditional on $\mathcal{H}_{t-1}$ and denote by $\pi^*_t(\theta^*)$ and $\pi^*_t(\theta^*\,|\,\tilde{\theta}^*_t)$ the marginal and conditional-on-$\theta^*_t$ density functions of $\theta^*$ given $\mathcal{H}_{t-1}$, noting that these random variables are constructed such that $\tilde{\pi}_t(\theta)=\tilde{\pi}^*_t(\theta)$ for all $\theta$ and, conditional on $\mathcal{H}_t$, $\tilde{\theta}_t$ is independent of both $\tilde{\theta}^*_t$ and $\theta^*$. We find that
\begin{align}
    &~ \mathbb{E}_{t-1}\left( \mu_t\left(\alpha_t(\tilde\theta_t),\theta^*\right)-\mu_t\left(\alpha_t(\tilde\theta^*_t),\theta^*\right) \right) \nonumber \\
    &= \sum_{\tilde{\theta}_t} \sum_{\tilde{\theta}^*_t} \int \left( \mu_t\left(\alpha_t(\tilde\theta_t),\theta^*\right)-\mu_t\left(\alpha_t(\tilde\theta^*_t),\theta^*\right) \right) \tilde{\pi}_t(\tilde{\theta}_t) \tilde{\pi}^*_t(\tilde{\theta}^*_t) \pi^*_t(\theta^*\,|\,\tilde{\theta}^*_t)\,{\rm d}\theta^*\nonumber \\
    &= \sum_{\tilde{\theta}_t} \int  \mu_t\left(\alpha_t(\tilde\theta_t),\theta^*\right) \tilde{\pi}_t(\tilde{\theta}_t) \pi^*_t(\theta^*)\,{\rm d}\theta^*
     - \sum_{\tilde{\theta}^*_t} \int \mu_t\left(\alpha_t(\tilde\theta^*_t),\theta^*\right) \tilde{\pi}^*_t(\tilde{\theta}^*_t) \pi^*_t(\theta^*\,|\,\tilde{\theta}^*_t)\,{\rm d}\theta^*\nonumber \\
     &= \sum_{\tilde{\theta}^*_t} \left\{ \int  \mu_t\left(\alpha_t(\tilde\theta^*_t),\theta^*\right) \pi^*_t(\theta^*)\,{\rm d}\theta^* - \int  \mu_t\left(\alpha_t(\tilde\theta^*_t),\theta^*\right) \pi^*_t(\theta^*\,|\,\tilde\theta^*_t)\,{\rm d}\theta^*\right\}\tilde{\pi}^*_t(\tilde\theta^*_t). \nonumber\\
     &= \sum_{\tilde{\theta}^*_t} \left\{ \int  \mu_t\left(\alpha_t(\tilde\theta^*_t),\theta^*\right) \left(\pi^*_t(\theta^*) -  \pi^*_t(\theta^*\,|\,\tilde\theta^*_t)\right)\,{\rm d}\theta^*\right\}\tilde{\pi}^*_t(\tilde\theta^*_t). \nonumber
\end{align}
Recall from the definition of \REV{$\tilde\Theta_t$} that when $\tilde{\theta}^*_t\in \REV{\tilde\Theta_t}$, the optimal response is $\alpha_t(\tilde{\theta}^*_t)=1$, and (from \eqref{eq::lossfun}) the loss function is $1-(l_{11}-1)\sigma(x_t^T\theta^*)$. When $\tilde{\theta}^*_t\in\Theta\setminus\REV{\tilde\Theta_t}$ the optimal action is $\alpha_t(\tilde{\theta}^*_t)=0$, and the loss function is $l_{01}\sigma(x_t^T\theta^*_t)$. 
Continuing the above derivation, and noting that $\int \pi^*_t(\theta^*)\mathrm{d}\theta^*=\int\pi_t^*(\theta^*|\tilde\theta^*_t)\mathrm{d}\theta^* = 1$), we have,
\begin{align}
    &~ \mathbb{E}_{t-1}\left( \mu_t\left(\alpha_t(\tilde\theta_t),\theta^*\right)-\mu_t\left(\alpha_t(\tilde\theta^*_t),\theta^*\right) \right) \nonumber \\
    &= \sum_{\tilde\theta_t^*\in\REV{\tilde\Theta_t}}\left\{\int \left(1+(l_{11}-1)\sigma(x_t^\top\theta^*)\right)\left(\pi_t^*(\theta^*)-\pi_t^*(\theta^*~|~\tilde\theta^*_t)\right)\mathrm{d}\theta^*\right\} \tilde\pi^*_{t}(\tilde\theta^*_t)\nonumber \\
    &\quad \quad +\sum_{\tilde\theta_t^*\in\Theta\setminus\REV{\tilde\Theta_t}}\left\{\int l_{01}\sigma(x_t^\top\theta^*)\left(\pi_t^*(\theta^*)-\pi_t^*(\theta^*~|~\tilde\theta^*_t)\right)\mathrm{d}\theta^*\right\}  \tilde\pi^*_{t}(\tilde\theta^*_t) \nonumber \\
    &= \sum_{\tilde\theta_t^*\in\REV{\tilde\Theta_t}}\left\{\int (l_{11}-1)\sigma(x_t^\top\theta^*)\left(\pi_t^*(\theta^*)-\pi_t^*(\theta^*~|~\tilde\theta^*_t)\right)\mathrm{d}\theta^*\right\} \tilde\pi^*_{t}(\tilde\theta^*_t)\nonumber \\
    &\quad \quad +\sum_{\tilde\theta_t^*\in\Theta\setminus\REV{\tilde\Theta_t}}\left\{\int l_{01}\sigma(x_t^\top\theta^*)\left(\pi_t^*(\theta^*)-\pi_t^*(\theta^*~|~\tilde\theta^*_t)\right)\mathrm{d}\theta^*\right\} \tilde\pi^*_{t}(\tilde\theta^*_t) \nonumber \\
    &\leq \max(l_{01},1-l_{11})\sum_{\tilde\theta_t^*}\left|\int\sigma(x_t^\top\theta^*)\pi_t^*(\theta^*)\mathrm{d}\theta^*-\int \sigma(x_t^\top\theta^*)\pi_t^*(\theta^*~|\tilde\theta^*_t)\mathrm{d}\theta^*\right| \tilde\pi^*_{t}(\tilde\theta^*_t). \label{eq::expandexp}
    \end{align}

Next, consider the denominator. We have \begin{align*}
    \REV{I_{t}}\left(\tilde\theta_t^*;~ (\tilde\theta_t,\Phi_t(\alpha_t(\tilde\theta_t)))\right) 
    &=  \REV{I_{t}}\left(\tilde\theta^*_t ;~ \tilde\theta_t \right) +   \REV{I_{t}}\left(\tilde\theta^*_t ;~ \Phi_t(\alpha_t(\tilde\theta_t)) ~|~  \tilde\theta_t \right) \\
    &=   \REV{I_{t}}\left(\tilde\theta^*_t ;~ \Phi_t(\alpha_t(\tilde\theta_t)) ~|~  \tilde\theta_t \right) \\
    &= \sum_{\theta}   \REV{I_{t}}\left(\tilde\theta^*_t ;~ \Phi_t(\alpha_t(\theta))  \right) \tilde\pi_{t}(\theta) \\
    &= \tilde\pi_{t}(\REV{\tilde\Theta_t})  \REV{I_{t}}\left(\tilde\theta^*_t;~ \Phi_t(1) \right),
\end{align*}
where we have used the conditional independence of $\tilde{\theta}_t$ and $\tilde{\theta}^*_t$ and the fact that the information gain is zero if $A_{t}=0$ is chosen. Then, rewriting the mutual information in terms of KL-divergence and applying Pinsker's inequality, we have the following lower bound, \begin{align}
    &~ ~ \tilde\pi_{t}(\REV{\tilde\Theta_t})  \REV{I_{t}}\left(\tilde\theta^*_t ;~ \Phi_t(1) \right) \nonumber \\
    &= \tilde\pi_{t}(\REV{\tilde\Theta_t}) \sum_{\tilde\theta_t^*} KL\left[\pi_{t-1}\left(\Phi_t(1)~|~\tilde\theta_t^*\right)~||~\pi_{t-1}\left(\Phi_t(1)\right) \right] \tilde\pi^*_t(\tilde\theta^*_t) \nonumber \\
    &\geq 2\tilde\pi_{t}(\REV{\tilde\Theta_t})  \sum_{\tilde\theta_t^*} d_{TV}\left(\pi_{t-1}\left(\Phi_t(1)~|~\tilde\theta_t^*\right),~\pi_{t-1}\left(\Phi_t(1)\right)  \right)^2 \tilde\pi^*_{t}(\tilde\theta^*_t) \nonumber \\ 
    &= 2\tilde\pi_{t}(\REV{\tilde\Theta_t})  \sum_{\tilde\theta_t^*} \left\{\int\sigma\left(x_t^\top\theta^*\right)\pi_t^*(\theta^*)\mathrm{d}\theta^*-\int\sigma\left(x_t^\top \theta^*\right)\pi_t^*(\theta^*~|~\tilde\theta^*_t)\mathrm{d}\theta^*\right\}^2  \tilde\pi^*_{t}(\tilde\theta^*_t).\label{eq::infgainlb}
\end{align}

Combining \eqref{eq::expandexp} and the lower bound \eqref{eq::infgainlb} we realise a bound on the information ratio, as below. We have, \begin{align*}
    \Gamma_t(\tilde\theta^*_t,\tilde\theta_t) &\leq \frac{\left(\max(l_{01},1-l_{11})\sum_{\tilde\theta_t^*}\left\{\int\sigma(x_t^\top\theta^*)\pi_t^*(\theta^*)\mathrm{d}\theta^*-\int \sigma(x_t^\top\theta^*)\pi_t^*(\theta^*~|\tilde\theta^*_t)\mathrm{d}\theta^*\right\} \tilde\pi^*_{t}(\tilde\theta^*_t)\right)^2}{2\tilde\pi_{t-1}(\Theta_t)\sum_{\tilde\theta_t^*} \left\{\int\sigma(x_t^\top\theta^*)\pi_t^*(\theta^*)\mathrm{d}\theta^*-\int\sigma(x_t^\top\theta^*)\pi_t^*(\theta^*~|~\tilde\theta^*_t)\mathrm{d}\theta^*\right\}^2  \tilde\pi^*_{t}(\tilde\theta^*_t)} \\
    &\leq \frac{\max(l_{01},1-l_{11})^2}{2\tilde\pi_{t-1}(\REV{\tilde\Theta_t})},
\end{align*} where the final inequality holds by Cauchy-Schwarz. $\square$

\section{Proof of Proposition \ref{prop::partitionsize}} \label{app::partitionsize} 

Note that for a given $\theta'$, and round $t \in [T]$ there are at most three values of the distortion rate. We have, for $\theta' \in \Theta$, and $t\in[T]$, \begin{equation*}
    d_{t}(\theta,\theta') = \begin{cases} 
                           &0, \quad \quad \quad \quad \quad \quad \quad \quad \quad \quad \quad \quad \text{ if } \alpha_t(\theta')=\alpha_t(\theta) \\
                           &(1+l_{01}-l_{11})\sigma(x_t^\top\theta')-1, ~\text{ if } \alpha_t(\theta')=0 \text{ and } \alpha_t(\theta)=1 \\
                           &1-(1+l_{01}-l_{11})\sigma(x_t^\top\theta'), ~\text{ if } \alpha_t(\theta')=1 \text{ and } \alpha_t(\theta)=0.
    \end{cases}
\end{equation*}

The condition \eqref{eq::partcond} is equivalent to \begin{equation*}
    \max_{t\in[T]}d_t(\theta,\theta')\leq \epsilon, ~\theta,\theta' \in \Theta_k, ~ \forall k \in [K].
\end{equation*} Since $d_t$ depends on the round $t$ only through $x_t$, we may replace the above condition with the following, \begin{equation}
    \label{eq::partX}
    d_\mathcal{X}(\theta,\theta') := \max_{x\in\mathcal{X}}d(\theta,\theta';x)\leq \epsilon, ~\theta,\theta' \in \Theta_k,~ \forall k \in [K].
\end{equation} Then any partition satisfying \eqref{eq::partX} satisfies condition \eqref{eq::partcond} for all $T$.

To identify the size of such a partition we consider the form of $d_\mathcal{X}$. Recall that $d(\theta,\theta';x)\neq 0$ only when $\alpha(\theta;x)\neq \alpha(\theta';x)$, thus the context vector $x\in\mathcal{X}$ achieving $\max_{x\in\mathcal{X}}d(\theta,\theta';x)$ must be such that $\theta$ and $\theta'$ select different classifications.\footnote{The only exception is in trivial settings where $x$ is so extreme that there is a single optimal action for all $\theta\in\Theta$.} Thus, we may write \begin{align*}
    d_\mathcal{X}(\theta,\theta') := \max_{x\in\mathcal{X}:\alpha(\theta;x)\neq\alpha(\theta';x)}d(\theta,\theta';x)= \max_{x\in\mathcal{X}:\alpha(\theta;x)\neq\alpha(\theta';x)}\left|1-(1+l_{01}-l_{11})\sigma\left(x^\top\theta'\right) \right|.
\end{align*}

For a parameter vector $\theta\in\Theta$, define the set of contexts on the classification boundary, $\mathcal{X}_\theta\subset \mathcal{X}$, as \begin{equation*}
    \mathcal{X}_\theta = \left\lbrace x \in \mathcal{X}: x^\top\theta = \log\left(\frac{1}{l_{01}-l_{11}}\right) \right\rbrace,
\end{equation*} (note: $\sigma^{-1}\left(1/({1+l_{01}-l_{11})}\right)=\log(1/(l_{01}-l_{11}))$). We then have for $\theta\in\Theta$ fixed that \begin{equation*}
    d_\mathcal{X}(\theta,\theta')\leq d_{\mathcal{X}_\theta}(\theta,\theta') := \max_{x\in\mathcal{X}_\theta}\left|1-(1+l_{01}-l_{11})\sigma\left(x^\top\theta'\right) \right|.
\end{equation*} It follows that $d_\mathcal{X}(\theta,\theta')\leq \epsilon$ for all $\theta'$ such that for all $x\in\mathcal{X}_\theta$ \begin{equation*}
    x^\top\theta'\in \left[\sigma^{-1}\left(\frac{1-\epsilon}{1+l_{01}-l_{11}}\right), \sigma^{-1}\left(\frac{1+\epsilon}{1+l_{01}-l_{11}}\right)\right]=\left[\log\left(\frac{1-\epsilon}{l_{01}-l_{11}+\epsilon}\right), \log\left(\frac{1+\epsilon}{l_{01}-l_{11}-\epsilon}\right)\right],
\end{equation*} i.e. over the same range that $d_{\mathcal{X}_\theta}(\theta,\theta')\leq\epsilon$. Expressing this in terms of the $x$-weighted norm between $\theta$ and $\theta'$ we can equivalently say that, $d_\mathcal{X}(\theta,\theta')\leq \epsilon$ for all $\theta'$ such that for all $x\in\mathcal{X}_\theta$, \begin{align*}
    x^\top(\theta - \theta')  &\in \left[\log\left(\frac{(l_{01}-l_{11})-\epsilon}{(l_{01}-l_{11})+(l_{01}-l_{11})\epsilon}\right), \log\left(\frac{(l_{01}-l_{11})+\epsilon}{(l_{01}-l_{11})-(l_{01}-l_{11})\epsilon}\right) \right] \\
    &\quad = \left[\log\left(\frac{1-\frac{\epsilon}{l_{01}-l_{11}}}{1+\epsilon}\right), ~ \log\left(\frac{1+\frac{\epsilon}{l_{01}-l_{11}}}{1-\epsilon}\right) \right],
    \end{align*} and thus also for all $\theta'$ such that for all $x \in \mathcal{X}_\theta$, \begin{align} \left|x^\top(\theta-\theta') \right| &\leq \min\left\lbrace \log\left(\frac{1+\frac{\epsilon}{l_{01}-l_{11}}}{1-\epsilon} \right),~-\log\left(\frac{1-\frac{\epsilon}{l_{01}-l_{11}}}{1+\epsilon} \right)\right\rbrace \nonumber \\
    &= \log\left(1+ \frac{\epsilon}{\min(1,l_{01}-l_{11})} \right)-\log\left(1-\frac{\epsilon}{\max(1,l_{01}-l_{11})}\right). \label{eq::xnormbound}
\end{align} The minimum can be shown to be defined as such by considering the difference  \begin{equation*}
    \log\left(\frac{1+\frac{\epsilon}{l_{01}-l_{11}}}{1-\epsilon} \right) - \left(- \log\left(\frac{1-\frac{\epsilon}{l_{01}-l_{11}}}{1+\epsilon} \right)\right) = \log\left(\frac{1-\frac{\epsilon^2}{(l_{01}-l_{11})^2}}{1-\epsilon^2} \right),
\end{equation*} and observing that it is negative when $(l_{01}-l_{11})<1$, for $\epsilon>0$.

We next move from the logarithmic bound \eqref{eq::xnormbound} to a bound that is linear in $\epsilon$. From the well-known logarithm inequalities, \begin{equation*}
    \frac{x}{1+x} < \log(1+x) < x, \quad x>1
\end{equation*} we also have for $a>0$ that, \begin{align*}
    \frac{ax}{1+ax} < \log(1+ax) &< ax, \quad x > -1/a, \\
    \text{ and } \frac{-ax}{1-ax} < \log(1-ax) &< -ax, \quad x < 1/a.
\end{align*} Defining $l_{min}=\min(l_{01},1-l_{11})$ and $l_{max}=\max(l_{01},1-l_{11})$, it follows that for $\epsilon\in [0,l_{max}^{-1})$, \begin{align}
    \log\left(1+ \frac{\epsilon}{l_{min}} \right)-\log\left(1-\frac{\epsilon}{l_{max}}\right) > \frac{\epsilon/l_{min}}{1+\epsilon/l_{min}} + \frac{\epsilon}{l_{max}} > \frac{\epsilon}{l_{max}}. \label{eq::logbound}
\end{align}

Thus by the combination of \eqref{eq::xnormbound} and \eqref{eq::logbound}, for a given $\theta \in \Theta$, we have that $d_\mathcal{X}(\theta,\theta')\leq \epsilon$ for all $\theta' \in\Theta$ such that \begin{equation*}
    \sum_{i=1}^d|\theta_i-\theta'_i| \leq \frac{\epsilon}{\max_{x\in\mathcal{X}_\theta}||x||\cdot l_{max}}.
\end{equation*} It follows that the size of a partition satisfying condition \eqref{eq::partcond} may be bounded by the size of an $\epsilon/(l_{max}\max_{x\in\mathcal{X}}||x||)$-cover of $\Theta$ with respect to the $\ell_1$ norm. Since $\Theta\subset B^d_1$, the size of the cover of $\Theta$ is itself bounded by the size of the cover of $B^d_1$, and therefore we have,  \begin{equation*}
    K\leq \left(\frac{3l_{max}x_{max}}{\epsilon}\right)^d, 
\end{equation*} via the standard result (see e.g. Lemma 1 of \cite{lorentz1966metric}) that a $\delta$-cover of a unit ball in $d$ dimensions is of size $(3/\delta)^d.$ $\square$

\section{Polya-Gamma Gibbs Sampler} \label{app:gibbs}

Following $t$ rounds, where $\theta^*$ has prior $\pi$, the posterior distribution on $\theta^*$ is as follows, \begin{equation}
    \pi(\theta ~|~ \mathcal{H}_t) = \frac{\pi(\theta)}{D_t}\prod_{s \in [t]:A_s=1} \left(\sigma(x_s^\top \theta) \right)^{C_s} \left( 1-\sigma(x_s^\top \theta) \right)^{1-C_s}, ~\theta\in\Theta, \label{eq::posterior}
\end{equation} where $D_t$ is the normalising constant, \begin{equation} \label{eq::normconst}
    D_t = \int_\Theta \pi(\theta) \prod_{s \in [t]:A_s=1} \left(\sigma(x_s^\top \theta) \right)^{C_s} \left( 1-\sigma(x_s^\top \theta) \right)^{1-C_s} d\theta.
\end{equation} Regardless of the choice of prior $\pi$, the posterior in \eqref{eq::posterior} is intractable, in the sense that samples cannot readily be drawn directly from it. However, if $\pi$ is chosen as a multivariate Gaussian then a highly efficient approximate sampling scheme is achievable via augmentation of the likelihood with PG random variables. 

A real-valued random variable follows a PG distribution with parameters $b>0$ and $c\in\mathbb{R}$, $X\sim PG(b,c)$, if \begin{equation*}
    X=\frac{1}{2\pi^2}\sum_{k=1}^\infty\frac{G_k}{(k-\frac{1}{2})^2+\frac{c^2}{4\pi^2}},
\end{equation*} where $G_k$ are i.i.d. $Gamma(b,1)$ random variables. Key to the augmentation scheme is the following identity of \cite{PolsonEtAl2013}, for $a\in \mathbb{R}$, $b>0$ and $\omega$ following a $PG(b,0)$ distribution \begin{equation*}
    \frac{\left(e^z\right)^a}{\left(1+e^z\right)^b} = 2^{-b}e^{(a-b/2)z}\int_0^\infty e^{-\omega z^2/2}p(\omega)\mathrm{d}\omega, ~~ z \in \mathbb{R}.
\end{equation*} 

Applying this identity to the likelihood component of \eqref{eq::posterior}, we may write \begin{align*}
    \pi(\theta ~|~ \mathcal{H}_{t}) \propto {\pi(\theta)}\prod_{s\in[t]:A_s=1}\exp\left((1/2-C_s)x_s^\top\theta\right)\int_0^\infty \exp\left(-\omega_s(x_s^\top\theta)^2/2\right)p(\omega_s)\mathrm{d}\omega_s,
\end{align*} where each $\omega_s$ is a $PG(1,0)$ random variable. It follows that if $\pi(\theta)$ is chosen as a multivariate Gaussian, s.t. $\theta^*\sim MVN(\mathbf{b},\mathbf{B})$, the posterior on $\theta^*$, conditioned on PG random variables $\bs\omega=(\omega_1,\dots,\omega_{|...|})$, is also multivariate Gaussian, \begin{equation*}
    \pi(\theta ~|~\bs\omega,\mathcal{H}_{t}) \propto \pi(\theta)\prod_{s\in[t]:A_s=1} \exp\left(\frac{\omega_s}{2}\left(x_s^\top\theta-\frac{(1-C_s)}{\omega_s}\right)^2\right).
\end{equation*} As such, we may construct a Gibbs sampler, which iterates between sampling PG random variables, and from the conditional Gaussian on $\theta^*$. Sampling of PG random variables is highly efficient, due to a rejection sampler of \cite{PolsonEtAl2013} with acceptance probability no less than 0.9992.

Our PG-TS approach (including the Gibbs steps) is summarised in Algorithm \ref{alg::PGTS}. It uses an additional counter random variable $N(t)=\sum_{s=1}^t\mathbb{I}\{A_s=1\}$ to track the number of rounds in which the class $1$ has been chosen, and assumes (in line with the definition of the model) that $C_t$ can be recovered from $\Phi_t(1)$. Further, for $n\in\mathbb{N}$ it uses $R(n)=\min(t\geq n: N(t)=n)$ to refer to the round in which class 1 is chosen for the $n^{th}$ time, and $\mathbf{X}_n$ to represent the matrix whose columns are the feature vectors $x_{R(1)},x_{R(2)},\dots,x_{R(n)}$ (in that order).

In each round PG-TS draws $M$ via the Gibbs sampling possible due to PG augmentation. Here we describe the simplest version of the algorithm, in the sense that a fixed number of samples, $M$, are used to estimate the posterior in each round, but a time dependent $M(t)$ could also be used. 

\begin{algorithm}[htbp]
    \caption{GIBBS}
    \label{alg::PG-Gibbs}
    \hrule
    \vspace{0.2cm}
    \textbf{Inputs:} Prior mean vector $\mathbf{b}$, Prior precision matrix $\mathbf{B}$, Number of Gibbs iterations $M$, Observed classes $\mathbf{C}=(C_1,\dots,C_n)$, Context matrix $\mathbf{X}\in \mathbb{R}^{d\times n}$, Initialisation parameter $\theta^{(0)}$ \\
    \vspace{0.2cm} \hrule \vspace{0.2cm}
    Compute $\bs\kappa = (C_1-\frac{1}{2}, \dots,C_n - \frac{1}{2})$ \\ 
         \textbf{for} $m=1$ \textbf{to} $M$ \textbf{do} \\ 
         \qquad \textbf{for} $i=1$ \textbf{to} $n$  \textbf{do} \\ 
         \qquad \qquad Draw $\omega_i~|~\theta^{(m-1)} \sim PG(1, {x}_{i}^\top \theta^{(m-1)}).$ \\ 
         \qquad \textbf{end for} \\
         \qquad Compute $\Omega = \text{diag}(\omega_1,\dots,\omega_{n})$ \\ 
         \qquad Compute covariance matrix $V_{\bs\omega} = (\mathbf{X}^\top \Omega\mathbf{X}+\mathbf{B}^{-1})^{-1}$ \\ 
         \qquad Compute mean vector $m_{\bs\omega} = V_{\bs\omega}(\mathbf{X}^\top \bs\kappa + \mathbf{B}^{-1}\mathbf{b})$ \\ 
         \qquad Draw $\theta^{(m)}~|~ \bs\kappa,\bs\omega \sim MVN(m_{\bs\omega},V_{\bs\omega})$ \\ 
         \textbf{end for} \\
         \textbf{return} $\{\theta^{(1)},\dots,\theta^{(M)}\}$
         \vspace{0.2cm}
      \hrule
      \vspace{0.2cm}
\end{algorithm}

\section{Proof of Theorem \ref{thm::no_divergence}} \label{app::no_divergence}

The proof utilises uniform ergodicity of the Gibbs sampler, together with convergence of the posterior to demonstrate that the $\upalpha$-divergence is vanishing in the limit.

\cite{choi2013polya} have shown that the PG-Gibbs sampler is uniformly ergodic, and thus as the number of Gibbs samples $M$ approaches $\infty$, the distribution $\pi_t^{(M)}$ shows convergence to $\pi_t$. Specifically, that there exists $\rho\in[0,1)$ such that, \begin{equation*}
    \left|\pi_T-\pi_T^{(M)} \right|_{TV} \leq \left|\pi_T-\pi_T^{(0)}\right|_{TV}\rho^M,
\end{equation*} where $|\cdot|_{TV}$ denotes the total variation distance between probability distributions. Here, however, we consider an analysis of the finite $M$, infinite $T$ setting, so this result alone does not imply shrinkage of the $\upalpha$-divergence.

To prove such a result, we must show that TS using either of $\pi_t$ or $\pi_t^{(M)}$ will lead to using the informative action (i.e. $A_t=1$) infinitely often. In either case, the draws of samples $\theta_t$ from $\pi_{t-1}$ or $\theta_t^{(M)}$ from $\pi_{t-1}^{(M)}$ are independent of the draw of a context $x_t$ from $p_X$. As such, by Assumption \ref{assum:1} we have $\mathbb{E}_{t-1}(p_X(\mathcal{X}_1(\theta_t)))>\delta$ and $\mathbb{E}_{t-1}^{(M)}(p_X(\mathcal{X}_1(\theta_t^{(M)})))>\delta$, for all $t$. This provides the following guarantees on the rate of selection of the informative action, \begin{align}
\lim_{T\rightarrow\infty}\mathbb{E}_0\left(\frac{\sum_{t=1}^T\mathbb{I}\{A_t^{TS}=1\}}{T}\right) &>\delta, \text{ and} \label{eq::infsample} \\
\lim_{T\rightarrow\infty}\mathbb{E}_0\left(\frac{\sum_{t=1}^T\mathbb{I}\{A_t^{(M)}=1\}}{T}\right) &>\delta. \label{eq::infsampPG}
\end{align} 

It therefore follows that under either sampling regime (i.e. either the exact or approximate TS algorithms) the number of observed labels approaches infinity as $T$ does. As such, the posterior induced under either regime is strongly consistent \citep[Proposition 1]{ghosal1995convergence} and satisfies the following stationary convergence guarantee \citep[Lemma 3.7]{yang2013sequential} with respect to the $L_1$ norm: \begin{equation*}
    \left|\pi_T-\pi_{T-1}\right|_1 \rightarrow 0, ~\text{as } T\rightarrow \infty.
\end{equation*}

Now fix $\epsilon>0$. Since the TV norm is bounded by half the $L_1$ norm between densities (when the densities exist --- see e.g. equation (1) of \cite{devroye2018total}), there exists $S>0$ such that $|\pi_t-\pi_{t-1}|_{TV}<\epsilon(1-\rho^M)/(3\rho^M)$ for all $t>S$. Hence, we have, by repeated application of the triangle inequality and the uniform ergodicity result, that,
\begin{align*}
    |\pi_T-\pi_T^{(M)}|_{TV} &\leq |\pi_T-\pi_{T}^{(0)}|_{TV}\rho^M \\
    &\leq |\pi_T-\pi_{T-1}|_{TV}\rho^M + |\pi_{T-1}-\pi_T^{(0)}|_{TV}\rho^M \\
    &= |\pi_T-\pi_{T-1}|_{TV}\rho^M + |\pi_{T-1}-\pi_{T-1}^{(M)}|_{TV}\rho^M \\
    &\leq \sum_{t=1}^T |\pi_{T+1-t}-\pi_{T-t}|_{TV}\rho^{tM} + |\pi_{0}-\pi_{0}^{(M)}|_{TV}\rho^{MT}\\
    &= \sum_{t=1}^{T-S} |\pi_{T+1-t}-\pi_{T-t}|_{TV}\rho^{tM} \\
    &\quad \quad \quad \quad + \sum_{t=T-S+1}^T |\pi_{T+1-t}-\pi_{T-t}|_{TV}\rho^{tM}+ |\pi_{0}-\pi_{0}^{(M)}|_{TV}\rho^{MT}\\
    &< \frac{\epsilon(1-\rho^M)}{3\rho^M}\sum_{t=1}^{T-S} \rho^{tM} + \sum_{t=T-S+1}^T\rho^{tM} + \rho^{MT}\\
    &< \frac{\epsilon(1-\rho^M)}{3\rho^M}\sum_{t=1}^{\infty}\rho^{tM} + \rho^{(T-S)M}\sum_{t=1}^\infty\rho^{tM} + \rho^{TM}\\
    &= \frac{\epsilon}3 + \rho^{(T-S)M}\frac{\rho^M}{1-\rho^M} + \rho^{TM}
\end{align*}
For $T$ sufficiently large that $\rho^{(T-S)M}<(1-\rho^M)\epsilon/(3\rho^M)$ and $\rho^{TM}<\epsilon/3$, we see that $|\pi_T-\pi_T^{(M)}|_{TV}<\epsilon$. Since $\epsilon>0$ is arbitrary, we see that $|\pi_T-\pi_T^{(M)}|_{TV}\to 0$ as $T\to\infty$. $\square$

\section{Proof of Theorem \ref{thm::maylike}} \label{app::maylike}

The proof in this section is somewhat informal, but constructed as such with the intention of limiting simple but lengthy translations of existing results to near-identical settings. Ultimately, the proof amounts to demonstrating that since PG-TS will select the informative action infinitely often, and since the posterior approximation will converge to the underlying true parameter $\theta^*$, the proportion of rounds in which PG-TS selects the action which is optimal in expectation approaches 1 as the number of rounds approaches infinity.

A similar result is established in an alternative, and more traditional, contextual bandit setting by \cite{MayEtAl2012}, in their Theorem 1. Therein each action $a\in\mathcal{A}$ (where $|\mathcal{A}|$ may be greater than 2) is associated with a continuous reward function $f_a:\mathcal{X}\rightarrow \mathbb{R}$, and each of these functions may have separate parameters. \cite{MayEtAl2012} demonstrate that an asymptotic consistency result is enjoyed by any TS-like algorithm for such a problem subject to conditions on the sampling distribution. Therein a TS-like algorithm is defined as one which samples reward functions $\tilde{f}_{t,a}$ from distributions $Q_{t,a}$ at each time $t$, and plays the action with the largest $\tilde{f}_{t,a}(x_t)$ value, and sufficient conditions are expressed in terms of the distributions $Q_{t,a}$. 

The basis of our informal proof in this section is to demonstrate that LCAT is sufficiently similar to the aforementioned contextual bandit problem, and the approximate posterior distributions used by PG-TS satisfy appropriate conditions such that the asymptotic consistency result can be extended to this setting.

Two conditions are critical to the asymptotic consistency result in \cite{MayEtAl2012}. First, where $n_{t,a}=\sum_{s=1}^t\mathbb{I}\{A_s=a\}$ is defined to be the number of plays of action $a$ in $t$ rounds, that we have \begin{equation}
    \mathbb{P}\left( \cup_{a\in\mathcal{A}}\left\lbrace n_{t,a}\rightarrow \infty \text{ as } t\rightarrow \infty\right\rbrace \right), \label{eq::conscond1}
\end{equation} i.e. that every action is sampled infinitely often in the limit. Second, that for each action $a \in \mathcal{A}$, we have convergence of the sampling distribution to the true reward function, i.e. \begin{equation}
    \left[Q_{t,a}-f_a(x_t) \right] \rightarrow^{\mathbb{P}} 0 \text{ as } n_{t,a}\rightarrow \infty. \label{eq::conscond2}
\end{equation} These conditions are the ultimate (and critical) consequence of Assumptions 1, 2, and 4, and the intermediate Lemma 2 in \cite{MayEtAl2012}.

Proceeding, we first show that the LCAT problem may equivalently be viewed as a contextual bandit problem with $\mathcal{A}=2$, and a known reward function for one action. Although this was a step we avoided in the main Bayesian regret analysis, since a bespoke analysis was more powerful than relying on generic contextual bandit results, it is nevertheless useful in this instance, to obtain the consistency result for the approximate algorithm.

We recall the form of the loss and signal matrices used previously, \begin{equation*}
      \mathbf{L} =  \left( \begin{matrix} 0 & l_{01} \\ 1 & l_{11} \end{matrix}\right) \text{ and } \bs\Phi =  \left( \begin{matrix} 0 & 0 \\ 1 & l_{11} \end{matrix}\right),
\end{equation*} and define shifted versions of these, \begin{equation*}
      \mathbf{L}' =  \mathbf{L}-\left(\begin{matrix} 0 & l_{01} \\
      0 & l_{01}
      \end{matrix}\right)=\left( \begin{matrix} 0 & 0 \\ 1 & l_{11}-l_{01} \end{matrix}\right) \text{ and } \bs\Phi' =  \left( \begin{matrix} 0 & 0 \\ 1 & l_{11}-l_{01} \end{matrix}\right).
\end{equation*} The shifted loss matrix has merely changed the scale of the losses for the event $C_t=1$, and the shifted signal matrix replaces one (essentially) arbitrary signal $l_{11}$ with another $l_{11}-l_{01}$. 

The contextual partial monitoring problem characterised by loss matrix $\mathbf{L}'$ and signal matrix $\bs\Phi'$ is recognised as a 2-armed logistic contextual bandit. In particular, taking rewards to be the negative of losses, the action indexed 0 has expected reward function $$f_0(x)=0, ~~x \in \mathcal{X}$$ and the action indexed 1 has expected reward function $$f_1(x)=(1+l_{01}-l_{11})\sigma(x^\top\theta)-1, ~~ x \in\mathcal{X}.$$ The reward observations have zero noise under selection of action 0, and are supported on $\{-1,l_{01}-l_{11}\}$ under selection of action 1. A straightforward adaptation of the action selection step implements a version of PG-TS for this version of the problem using the same Gibbs sampler and structure.

Having framed the LCAT problem as a contextual bandit problem, proof of the asymptotic consistency result then reduces to the verification of conditions \eqref{eq::conscond1} and \eqref{eq::conscond2}. Firstly, we have from  \eqref{eq::infsampPG} that PG-TS will sample the informative action infinitely often, and by an equivalent proof under Assumption \ref{assum:1} the same is true of the non-informative action. Thus we have that \eqref{eq::conscond1} is satisfied by PG-TS. Plainly, \eqref{eq::conscond2} holds for the non-informative action since its rescaled reward function is known by design, and \eqref{eq::conscond2} is shown to hold for the informative action by the consistent estimation result of Theorem \ref{thm::no_divergence}. Thus, by extension of the results of \cite{MayEtAl2012}, we have the asymptotic consistency and asymptotically sublinear regret of the PG-TS algorithm. $\square$

\section{Pseudocode for {CBP-SIDE} \REV{and SupLogistic Algorithms} for Apple Tasting} \label{app:CBP}

In this section we provide the particular version of the more general {CBP-SIDE} \REV{and SupLogistic algorithms} used for the contextual logistic apple tasting problem. Due to the small action set, and specific loss model, the statement of \REV{these algorithms} can be streamlined. That said, \REV{for CBP-SIDE} the correct choice of estimator, confidence set, and various constants is still non-trivial. We explain what we believe to be the best theoretically-supported choice of these components \REV{in subsection \ref{sec::cbp}, and outline the SupLogistic algorithm in subsection \ref{sec::suplog}.}

\subsection{CBP-SIDE} \label{sec::cbp}

CBP-SIDE requires the identification of \emph{observer vectors}, $v_{ij}$ and $v_{ji}$, for each pair of actions, $i,j$ (in our case, the only action pairs are of course $\{0,0\}$, $\{0,1\}$, and $\{1,1\}$), which satisfy, \begin{equation*}
    l_i-l_j = S_i^\top v_{ij} - S_j^\top v_{ji},
\end{equation*} where $l_i$ and $l_j$ are columns of the loss matrix $L$ and $S_i$ and $S_j$ are \emph{signal matrices} - which are exactly the incidence matrices of the symbols in columns of $\Phi$. In our setting, we have $l_0=(0~ 1)$, $l_1= (l_{01}~ l_{11})$ and signal matrices, \begin{equation*}
    S_0 = \left(\begin{matrix} 1 ~ 1 \end{matrix}\right), \quad S_1=\left(\begin{matrix} 1 & 0 \\ 0 & 1 \end{matrix}\right).
\end{equation*} It is clear that valid $v_{ij}$ and $v_{ji}$ are not unique, but \cite{BartokSzepesvari2012} note that a theoretically optimal choice, is \begin{equation*}
    \left(\begin{matrix} v_{ij} \\
    -v_{ji} \end{matrix} \right) =  \left(\begin{matrix} S_i^\top ~ S_j^\top \end{matrix}\right)^+ (l_i-l_j).
\end{equation*} So in our case we may choose scalar $v_{01}=-l_{01}$ and $v_{10}=l_{11}-1$, and $v_{00}=v_{11}=0$.

In each round, the general CBP-SIDE algorithm computes a estimate $\hat\theta$ of the unknown parameter, and from this a regret term $\Delta_{ij}$ and a confidence term $c_{ij}$ for each pair of actions $i,j$, of the form \begin{align*}
    \Delta_{ij} &= v_{ij}\hat{q}_i + v_{ji}\hat{q}_j \\
    c_{ij}      &= |v_{ij}|w(i,t) +|v_{ji}|w(j,t),
\end{align*} where $\hat{q}_k$ is the estimated probability action $k$ is suboptimal and $w(\cdot,\cdot)$ is a confidence width function chosen to establish theoretical guarantees. Since we choose $v_{00}=v_{11}$ we can focus solely on the case $\{i,j\}=\{0,1\}$ and compute a single regret term \begin{equation*}
    \hat\Delta:=\Delta_{01}=-l_{01}\left(1-\sigma\left(x_t^\top\hat\theta\right)\right) + (l_{11}-1)\sigma\left(x_t^\top\hat\theta\right),
\end{equation*} and confidence term \begin{equation*}
    \hat{c}:=c_{01}=\left(1+l_{01}-l_{11}\right)w(t),
\end{equation*} with $w$ specified below.

We use the maximum likelihood estimator, projected back in to $\Theta$ as our estimator of $\theta^*$. Specifically, with respect to the matrix \begin{equation*}
    V_t = \sum_{s=1}^{t-1}\mathbb{I}\{A_t=1\}x_s^\top x_s,
\end{equation*} and the log-likelihood maximised over $\mathbb{R}^d$ at $\theta^{MLE}_t$, we define the estimator \begin{equation*}
    \hat\theta_t = \argmin_{\theta\in\Theta}||\theta-\theta^{MLE}||_{V_t^{-1}}^2.
\end{equation*}

We adapt the confidence width function given in \cite{BartokSzepesvari2012} for online multinomial logistic regression, leading to the following expression derived from the self-normalised inequalities of \cite{abbasi2011improved}, \begin{equation*}
    w(t)=C\left(\sqrt{2d(1+N(t-1)R^2/d) +2\log(1/\delta_{N(t-1)})}+Rd\right)\sqrt{x_t^\top V_t^{-1}x_t}.
\end{equation*} Here, the $\Theta-$ and $\mathcal{X}-$dependent constant $C=[\inf_{\theta\in\Theta,x\in\mathcal{X}} (1-\sigma(x^\top\theta))\sigma(x^\top\theta)]^{-1}$ arises from Lemma 3 of \cite{BartokSzepesvari2012}, $R$ is a bound on the 2-norm of the features $x\in\mathcal{X}$, $N(t)=\sum_{s=1}^t\mathbb{I}\{A_s=1\}$ counts the number of uses of action 1 in $t$ rounds (i.e. the size of the observed data), and $\delta_s=s^{-2}$ is chosen to realise an optimal regret bound. 

Finally, the action selection process also simplifies with respect to the general case, and we assign class 1, unless the regret term $\hat\Delta$ falls below $-\hat{c}$, indicating that a prediction of class 0 with sufficient confidence to avoid pass on the information gaining action. Algorithm \ref{alg::CBPSIDE} gives the adaptation of CBP-SIDE to apple tasting, incorporating the above specifications.

\begin{algorithm}[htbp]
    \caption{CBP-SIDE for Apple Tasting}
    \label{alg::CBPSIDE}
    \hrule
    \vspace{0.2cm}    
        \textbf{Inputs:} Loss parameters $l_{01},l_{11}$\\
        
        \textbf{Initialise:}  $\mathcal{D}=\emptyset$.\\
        
        \textbf{for} $t=1,2,\dots$ \textbf{do} \\
         \qquad Receive context ${x}_t\in \mathbb{R}^d$ \\  
         \qquad Compute regret estimate $\hat\Delta = (1+l_{01}-l_{11})\sigma(x_t\hat\theta_t)-l_{01}$\\
         \qquad Compute confidence width $\hat{c} = (1+l_{01}-l_{11})w(t)$ \\
         \qquad Select action $A_t = 1-\mathbb{I}\{\hat\Delta\leq-\hat{c} \}$ \\
         \qquad \textbf{if} $A_t =1$ \textbf{do} \\ 
         \qquad \qquad Observe $\Phi_t(1)\in\{1,l_{11}\}$ \\ 
         \qquad \qquad Augment $\mathcal{D} \leftarrow \mathcal{D} \cup \{{x}_t,\Phi_t(1)\}$ \\ 
         \qquad \textbf{end if} \\ 
         \textbf{end for}
         \vspace{0.2cm}
      \hrule
      \vspace{0.2cm}
\end{algorithm}

\subsection{\REV{SupLogistic}} \label{sec::suplog}

\REV{SupLogistic requires some modification for application to apple tasting, as it contains phases where actions are pulled randomly for explorations' sake, and where different confidence widths are constructed based on the context associated with each arm. Since the no-information action has no value in terms of exploration, we adapt the algorithm's `burn-in phase' to use only the informative action. Later, the computation of confidence widths can be simplified somewhat, since the context associated with one arm is the negative of that associated with the other - this symmetry meaning we require only one confidence term per bucket per round, rather than one per action per bucket per round. Otherwise we only adapt the algorithm as presented in \cite{jun2021improved} to use notation consistent with the rest of the paper, and include the full pseudocode in Algorithm \ref{alg::SupLog}.}

\begin{algorithm}[htbp]
    \caption{SupLogistic for Apple Tasting}
    \label{alg::SupLog}
    \REV{\hrule
    \vspace{0.2cm}    
        \textbf{Inputs:} Loss parameters $l_{01},l_{11}$, Exploration rate $\alpha$, Burn-in length $\tau$,  Horizon $T$.\\
        
        \textbf{Initialise:}  $\mathcal{D}=\emptyset$, $S=\lfloor \log_2(T) \rfloor$, and $\Psi_1=\Psi_2=\dots=\Psi_{S+1}=\emptyset$.\\
        
        \textbf{for} $t=1,2,\dots\tau$ \textbf{do} \\
         \qquad Receive context ${x}_t\in \mathbb{R}^d$ \\
         \qquad Select action $A_t=1$ \\
         \qquad Observe $\Phi_t(1)\in\{1,l_{11}\}$ \\ 
         \qquad Augment $\mathcal{D} \leftarrow \mathcal{D} \cup \{{x}_t,\Phi_t(1)\}$ \\
         \qquad Augment $\Psi_{(t-1) \text{ mod } (S+1) +1} \leftarrow \Psi_{(t-1) \text{ mod } (S+1) +1} \cup \{t\}$ \\
        \textbf{end for} \\
        \textbf{Initialise:} $\Psi_*=\Psi_{S+1}$, $\Psi_0=\emptyset$ \\
        \textbf{for} $t=\tau+1, \tau+2,\dots,T$ \textbf{do} \\
         \qquad Receive context ${x}_t\in \mathbb{R}^d$ \\
         \qquad Compute $\hat\theta_{t}^{\Psi_*}=$ as the MLE over data from rounds in $\Psi_*$ \\
         \qquad \textbf{Initialise:} $s=1$, $A_t=-1$ \\
         \qquad \textbf{while } $A_t=-1$ \textbf{ do} \\
         \qquad \qquad Compute $\hat\theta^{(s)}_t$ as the MLE over data from rounds in $\Psi_s$\\
         \qquad \qquad Compute bucket $s$ loss estimates $$l_{t,0}^{(s)}=l_{01}\left(1-\sigma\left(x_t^\top\theta_t^{(s)}\right)\right) \text{ and, } l_{t,1}^{(s)}=1+(l_{11}-1)\sigma\left(x_t^\top\theta_t^{(s)}\right)$$ \\
         \qquad \qquad Compute bucket $s$ Fisher matrix $H^{(s)}(\hat\theta_t^{\Psi_*})=\sum_{u\in\Psi_s}\sigma'(x_u^\top\hat\theta_t^{\Psi_*})x_u x_u^\top$ \\
         \qquad \qquad Compute bucket $s$ confidence width $w_t^{(s)} = \alpha\sqrt{2.2}||x_t||_{H^{(s)}(\hat\theta_t^{\Psi_*})^{-1}}$ \\
         \qquad \qquad \textbf{if } $w_t^{(s)} \geq 2^{-s}$ \textbf{ do} \\
         \qquad \qquad \qquad Select action $A_t = 1$ \\
         \qquad \qquad \qquad Augment $\Psi_s \leftarrow \Psi_s \cup \{t\}$ \\
         \qquad \qquad \textbf{else if } $w_t^{(s)} \leq T^{-1/2}$ \textbf{ do} \\
         \qquad \qquad \qquad Select action $A_t = \argmin_{a\in\{0,1\}} l_{t,a}^{(s)}$ \\
         \qquad \qquad \qquad Augment $\Psi_0 \leftarrow \Psi_0 \cup \{t\}$ \\
         \qquad \qquad \textbf{else do} \\
         \qquad \qquad \qquad Set $s \leftarrow s+1$ \\
         \qquad \qquad \textbf{end else} \\
         \qquad \textbf{end while} \\
         \qquad \textbf{if} $A_t =1$ \textbf{do} \\ 
         \qquad \qquad Observe $\Phi_t(1)\in\{1,l_{11}\}$ \\ 
         \qquad \qquad Augment $\mathcal{D} \leftarrow \mathcal{D} \cup \{{x}_t,\Phi_t(1)\}$ \\ 
         \qquad \textbf{end if} \\ 
         \textbf{end for}
         \vspace{0.2cm}
      \hrule
      \vspace{0.2cm}}
\end{algorithm}

\section{Parameter Tuning for PG-IDS} \label{app::parameter}
In this section we give the results of further experiments used to estimate the optimal parameter $\lambda$ for the PG-IDS scheme. We consider ten choices of $\lambda=\{0,0.05,\dots,0.45\}$ and compare the performance of the resulting PG-IDS algorithms with each other and which traditional PG-IDs and PG-TS (with the same prior) for comparison's sake. To identify a robust choice of parameter we investigate Problem \REV{(i)} and from the main text, but with a range of horizons, loss matrices and context distributions, \REV{and Problem (iii) to illustrate robustness in the setting where other algorithms may struggle}.

\REV{Recall that the base version of Problem \REV{(i)} has $d=5$ with each dimension of $\theta^*$ drawn uniformly on $[-1,1]$ and contexts drawn from a zero-mean multivariate Gaussian with identity covariance matrix. We construct different variants of Problem \REV{(i)} by varying $d, l_{01}$, and the context distribution}, as well as the problem horizon, $T$. We  keep $l_{11}=0.05$ fixed. \REV{Specifically, we trial each candidate algorithm (or parameterisation) over $50$ replications, where in each new replication, $d$ is drawn uniformly from the set $\{3,5,10,20\}$, $\theta$ is drawn uniformly from $[-1,1]^d$, a variance term $s$ is drawn from a $Gamma(1,1)$, all contexts are drawn from a zero-mean multivariate Gaussian with covariance matrix $sI_d$, and the loss $l_{01}$ is drawn uniformly from $[0.5,0.95]$.}

\begin{figure}
\centering
\begin{subfigure}{0.75\textwidth}
    \includegraphics[width=\textwidth]{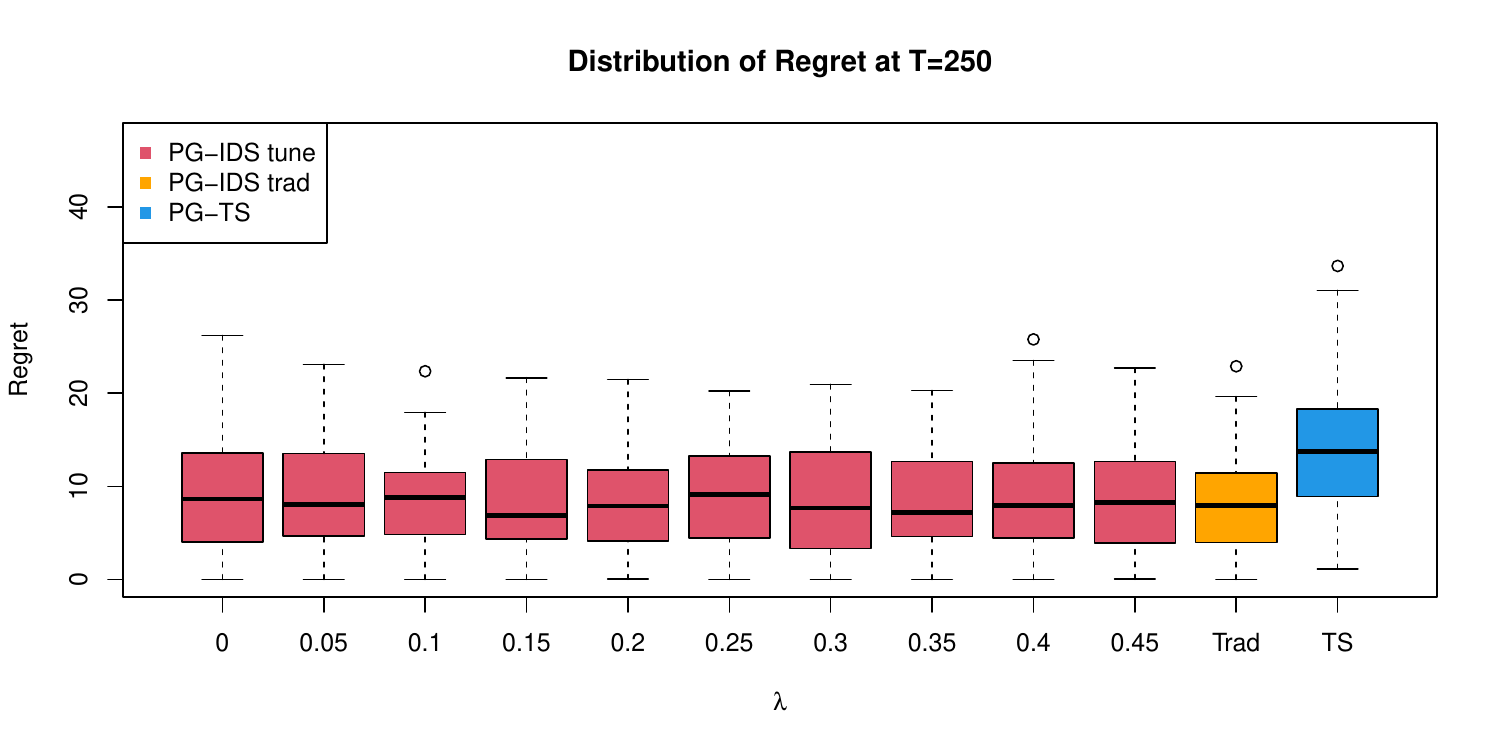}
    \caption{\REV{Problem (i) style experiments with $T=250$}}
\end{subfigure}
\begin{subfigure}{0.75\textwidth}
    \includegraphics[width=\textwidth]{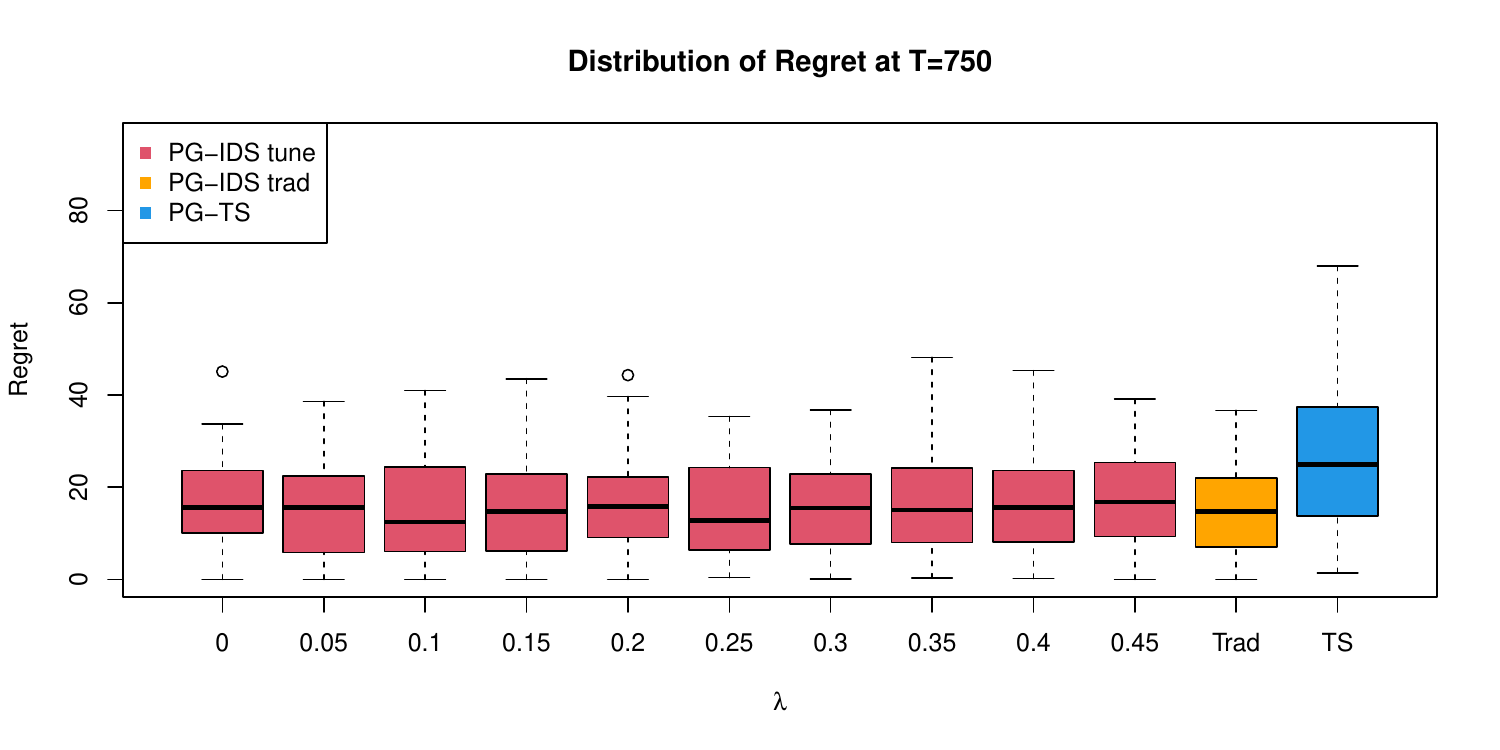}
    \caption{\REV{Problem (i) style experiments with $T=750$}}
\end{subfigure}
\begin{subfigure}{0.75\textwidth}
    \includegraphics[width=\textwidth]{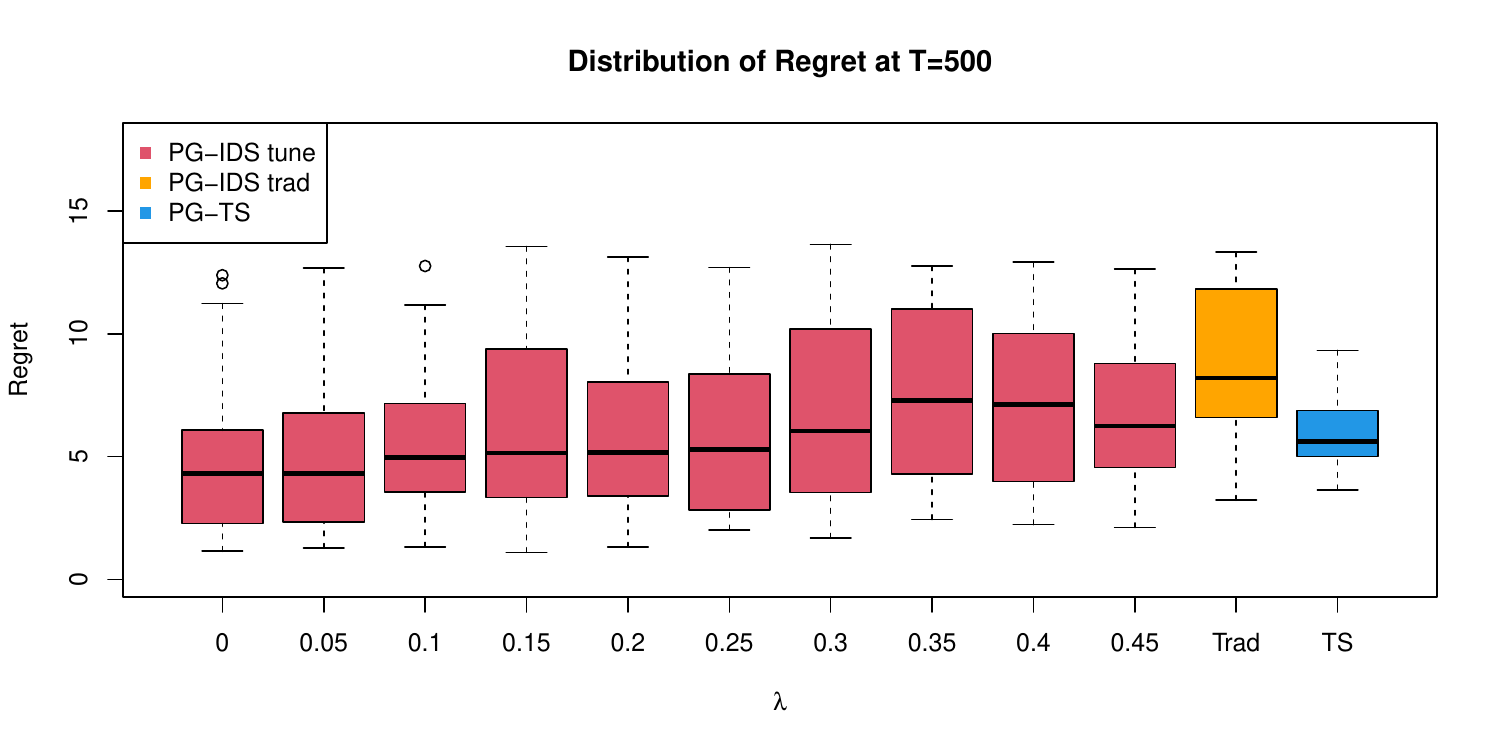}
    \caption{\REV{Problem (iii) experiments with $T=500$}}
\end{subfigure}
    \caption{\REV{Boxplots of the distribution of regret at final round for tunable PG-IDS with varying choices of tuning parameter $\lambda$ (in red), plotted alongside equivalent distributions for traditional PG-IDS (yellow) and TS (blue).}}
    \label{fig:tuneIDS1}
\end{figure}

\REV{The boxplots in Figure \ref{fig:tuneIDS1} (a) and (b), and statistics in Tables \ref{tab::250} and \ref{tab::750} summarise the performance of tunable PG-IDS with different values of $\lambda$, across the different variants of Problem (i). We see that performance is robust to the choice of $\lambda$ with little change in the distributions of regret incurred at rounds 250 and 750 across the values investigated. For Problem (iii), Figure \ref{fig:tuneIDS1} (c) and Table \ref{tab::Prob3Tune} provide the equivalent results, and a similar robustness. Overall $\lambda=0.05$ has the lowest mean regret combined across the three experiments, and is thus used in the comparisons in the main text. The experiments here give confidence that performance is fairly robust to this choice.}

\begin{table}[ht]
\centering
\begin{tabular}{rrrrrrrrrrr}
  \hline
  & \multicolumn{10}{c}{PG-IDS Tuning Parameter, $\lambda$}  \\
 & 0 & 0.05 & 0.10 & 0.15 & 0.20 & 0.25 & 0.30 & 0.35 & 0.40 & 0.45 \\
  \hline
Mean Regret & 9.70 & 9.30 & 8.86 & 8.60 & 8.77 & 9.01 & 8.69 & 8.83 & 9.20 & 8.73  \\ 
  St. Dev. & 6.66 & 6.15 & 4.93 & 5.67 & 5.76 & 5.45 & 5.64 & 5.39 & 6.17 & 5.52  \\ 
   \hline
\end{tabular}
\caption{\REV{Mean and standard deviations of regret at $T=250$ as $\lambda$ varies for tunable PG-IDS on Problem (i) variants.}}
\label{tab::250}
\end{table}
\begin{table}[ht]
\centering
\begin{tabular}{rrrrrrrrrrr}
  \hline
  & \multicolumn{10}{c}{PG-IDS Tuning Parameter, $\lambda$}  \\
 & 0 & 0.05 & 0.10 & 0.15 & 0.20 & 0.25 & 0.30 & 0.35 & 0.40 & 0.45 \\ 
  \hline
Mean Regret & 16.79 & 15.65 & 15.79 & 16.29 & 16.53 & 15.48 & 15.81 & 16.23 & 16.64 & 17.74  \\ 
  St. Dev. & 10.22 & 10.77 & 10.95 & 12.11 & 10.67 & 10.39 & 9.69 & 10.65 & 11.03 & 10.38  \\ 
   \hline
\end{tabular}
\caption{\REV{Mean and standard deviations of regret at $T=750$ as $\lambda$ varies for tunable PG-IDS on Problem (i) variants.}}
\label{tab::750}
\end{table}
\begin{table}[ht]
\centering
\begin{tabular}{rrrrrrrrrrr}
  \hline
  & \multicolumn{10}{c}{PG-IDS Tuning Parameter, $\lambda$}  \\
  & 0 & 0.05 & 0.10 & 0.15 & 0.20 & 0.25 & 0.30 & 0.35 & 0.40 & 0.45 \\ 
  \hline
Mean Regret & 4.81 & 5.01 & 5.62 & 6.26 & 6.10 & 5.79 & 6.87 & 7.42 & 7.38 & 6.82  \\ 
  St. Dev. & 2.97 & 3.17 & 2.94 & 3.62 & 3.24 & 3.33 & 3.63 & 3.39 & 3.35 & 3.05  \\ 
   \hline
\end{tabular}
\caption{\REV{Mean and standard deviations of regret at $T=500$ as $\lambda$ varies for tunable PG-IDS on Problem (iii).}}
\label{tab::Prob3Tune}
\end{table}

\end{document}